\documentclass[twocolumn,english]{IEEEtran}
\usepackage[T1]{fontenc}
\usepackage{color}
\usepackage{babel}
\usepackage{array}
\usepackage{amsmath}
\usepackage{amssymb}
\usepackage{graphicx}
\usepackage{epstopdf}
\usepackage{cite}
\usepackage[unicode=true,
 bookmarks=true,bookmarksnumbered=true,bookmarksopen=true,bookmarksopenlevel=1,
 breaklinks=false,pdfborder={0 0 0},pdfborderstyle={},backref=false,colorlinks=false]
 {hyperref}
\hypersetup{pdftitle={Your Title},
 pdfauthor={Your Name},
 pdfpagelayout=OneColumn, pdfnewwindow=true, pdfstartview=XYZ, plainpages=false}

\makeatletter

\providecommand{\tabularnewline}{\\}

 \let\oldforeign@language\foreign@language
 \DeclareRobustCommand{\foreign@language}[1]{%
   \lowercase{\oldforeign@language{#1}}}

\ifCLASSOPTIONcompsoc
\usepackage[caption=false,font=normalsize,labelfont=sf,textfont=sf]{subfig}
\else
\usepackage[caption=false,font=footnotesize]{subfig}
\fi

\@ifundefined{showcaptionsetup}{}{%
 \PassOptionsToPackage{caption=false}{subfig}}
\usepackage{subfig}
\makeatother

\begin{document}

\title{PALM: An Incremental Construction of Hyperplanes for Data Stream
Regression}

\author{Md Meftahul~Ferdaus,~\IEEEmembership{Student Member,~IEEE,} Mahardhika~Pratama,~\IEEEmembership{Member,~IEEE,} Sreenatha G.~Anavatti,~Matthew A.~Garratt,$\;$\thanks{Md Meftahul~Ferdaus, Sreenatha G. ~Anavatti, and Matthew A. Garratt
are with~the School of Engineering and Information Technology, University
of New South Wales at the Australian Defence Force Academy, Canberra,
ACT 2612, Australia (e-mail: \protect\href{mailto:m.ferdaus@student.unsw.edu.au}{m.ferdaus@student.unsw.edu.au};
\protect\href{mailto:s.anavatti@adfa.edu.au}{s.anavatti@adfa.edu.au};
\protect\href{mailto:M.Garratt@adfa.edu.au}{M.Garratt@adfa.edu.au}).}\thanks{Mahardhika Pratama is with the School
of Computer Science and Engineering, Nanyang Technological University,
Singapore, 639798, Singapore (e-mail: \protect\href{mailto:mpratama@ntu.edu.sg}{mpratama@ntu.edu.sg}). }}

\maketitle
\begin{abstract}
Data stream has been the underlying challenge in the age of big data because it calls for real-time data processing with the absence of a retraining process and/or an iterative learning approach. In realm of fuzzy system community, data stream is handled by algorithmic development of self-adaptive neuro-fuzzy systems (SANFS) characterized by the single-pass learning mode and the open structure property which enables effective handling of fast and rapidly changing natures of data streams. The underlying bottleneck of SANFSs lies in its design principle which involves a high number of free parameters (rule premise and rule consequent) to be adapted in the training process. This figure can even double in the case of type-2 fuzzy system. In this work, a novel SANFS, namely parsimonious learning machine (PALM),
is proposed. PALM features utilization of a new type of fuzzy rule based on the concept of hyperplane clustering which significantly reduces the number of network parameters because it has no rule premise parameters. PALM is proposed in both type-1 and type-2 fuzzy systems where all of which characterize a fully dynamic rule-based system. That is, it is capable of automatically generating, merging and tuning the hyperplane-based fuzzy rule in the single pass manner. Moreover, an extension of PALM, namely recurrent PALM (rPALM), is proposed and adopts the concept of teacher-forcing mechanism in the deep learning literature. The efficacy of PALM has been evaluated through numerical study with six real-world and synthetic data streams from public database and our own real-world project of autonomous vehicles. The proposed model showcases significant improvements in terms of computational complexity and number of required parameters against several renowned SANFSs, while attaining comparable and often better predictive accuracy.
\end{abstract}


\section{Introduction}

\IEEEPARstart{A}{dvance} in both hardware and software technologies has triggered generation of a large quantity of data in
an automated way. Such applications can be exemplified by space, autonomous
systems, aircraft, meteorological analysis, stock market analysis,
sensors networks, users of the internet, etc., where the generated
data are not only massive and possibly unbounded but also produced at a rapid
rate under complex environments. Such online data are known as data stream \cite{aggarwal2007data,gama2010knowledge}.
A data stream can be expressed in a more formal way \cite{silva2013data}
as $S=\left\{ x^{1},x^{2},...,x^{i},...,x^{\infty}\right\} ,$ where
$x^{i}$ is enormous sequence of data objects and possibly unbounded.
Each of the data object can be defined by an $n$ dimensional feature
vector as $x^{i}=[x_{j}^{i}]_{j=1}^{n},$ which may belong to a continuous,
categorical, or mixed feature space. In the field of data stream mining,
developing a learning algorithm as a universal approximator is challenging
due to the following factors 1) the whole data to train the learning
algorithm is not readily available since the data arrive continuously;
2) the size of a data stream is not bounded; 3) dealing with a huge
amount of data; 4) distribution of the incoming unseen data may slide
over time slowly, rapidly, abruptly, gradually, locally, globally,
cyclically or otherwise. Such variations in the data distribution
of data streams over time are known as $concept\;drift$ \cite{bose2014dealing,lughofer2011handling};
5) data are discarded after being processed to suppress memory consumption into practical level.

To cope with above stated challenges in data streams, the learning
machine should be equipped with the following features: 1) capability
of working in single pass mode; 2) handling various concept drifts in data streams; 3) has low memory burden and computational complexity to enable real-time deployment under resource constrained environment. In realm of fuzzy system, such learning aptitude is demonstrated by Self Adaptive Neuro-Fuzzy System (SANFS) \cite{juang1998online}. Until now, existing
SANFSs are usually constructed via hypersphere-based or hyperellipsoid-based clustering techniques (HSBC or HEBC) to automatically partition the input space into a number of fuzzy rule and rely on the assumption of normal distribution due to the use of Gaussian membership function \cite{angelov2004approach,angelov2005simpl_ets,pratama2014genefis,pratama2014panfis,pratama2017data,de2018error,pan2016hybrid,de2017usnfis,meda2018estimation}. As a result, they are always associated with rule premise parameters, the mean and width of Gaussian function, which need to be continuously adjusted. This issue complicates its implementation in a complex and deep structure. As a matter of fact, existing neuro-fuzzy systems can be seen as a single hidden layer feedforward network. Other than the HSSC or HESC, the data cloud based
clustering (DCBC) concept is utilized in \cite{angelov2012new,pratama2018parsimonious}
to construct the SANFS. Unlike the HSSC and HESC, the data clouds do
not have any specific shape. Therefore, required parameters in DCBC
are less than HSSC and HESC. However, in DCBC, parameters like mean,
accumulated distance of a specific point to all other points need
to be calculated. In other words, it does not offer significant reduction on the computational complexity and memory demand of SANFS. Hyperplane-Based Clustering (HPBC) provides a promising avenue to overcome this drawback because it bridges the rule premise and the rule consequent by means of the hyperplane construction.

Although the concept of HPBC already exists since the last two decades \cite{kim1997new,kung2007affine,li2009t}, all of them are characterized by a static structure and are
not compatible for data stream analytic due to their offline characteristics.
Besides, majority of these algorithms still use the Gaussian or bell-shaped Gaussian function \cite{zarandi2012type} to create the rule premise and are not free of the rule premise parameters. This problem is solved in \cite{zou2017ts}, where they
have proposed a new function to accommodate the hyperplanes directly in the rule premise.
Nevertheless, their model also exhibit a fixed structure and operates in the batch learning node. Based on
this research gap, a novel SANFS, namely parsimonious learning machine
(PALM), is proposed in this work.
The novelty of this work can be summarized as follows:
\begin{enumerate}
\item PALM is constructed using the HPBC technique and its fuzzy rule is fully characterized by a hyperplane which underpins both the rule consequent and the rule premise. This strategy reduces the rule base parameter to the level of $C*(P+1)$ where $C,P$ are respectively the number of fuzzy rule and input dimension.
\item PALM is proposed in both type-1 and type-2 versions derived from the concept of type-1 and type-2 fuzzy systems. Type-1 version incurs less network parameters and faster training speed than the type-2 version whereas type-2 version expands the degree of freedom of the type-1 version by applying the interval-valued concept leading to be more robust against uncertainty than the type-1 version.
\item PALM features a fully open network structure where its rules can be automatically generated, merged and updated on demand in the one-pass learning fashion. The rule generation process is based on the self-constructing clustering approach \cite{xu2015dimensionality,jiang2011fuzzy} checking coherence of input and output space. The rule merging scenario is driven by the similarity analysis via the distance and orientation of two hyperplanes. The online hyperplane tuning scenario is executed using the fuzzily weighted generalized recursive least square (FWGRLS) method.
\item an extension of PALM, namely recurrent PALM (rPALM), is put forward in this work. rPALM addresses the underlying bottleneck of HPBC method: dependency on target variable due to the definition of point-to-hyperplane distance \cite{pointPlane}. This concept is inspired by the teacher forcing mechanism in the deep learning literature where activation degree of a node is calculated with respect to predictor's previous output. The performance of rPALM has been numerically validated in our supplemental document where its performance is slightly inferior to PALM but still highly competitive to most prominent SANFSs in terms of accuracy.   
\item Two real-world problems from our own project, namely online identification of Quadcopter unmanned aerial vehicle (UAV) and helicopter UAV, are presented in this paper and exemplify real-world streaming data problems. The two datasets are collected from indoor flight tests in the UAV lab of the university of new south wales (UNSW), Canberra campus. These datasets, PALM and rPALM codes are made publicly available in \cite{palmcode}.
\end{enumerate}
The efficacy of both type-1 and type-2 PALMs have been numerically evaluated using six real-world and synthetic streaming data problems. Moreover, PALM is also compared against prominent SANFSs in the literature and demonstrates encouraging numerical results in which it generates compact and parsimonious network structure while delivering comparable and even better accuracy than other benchmarked algorithms.

The remainder of this paper is structured is as follows: Section~\ref{sec:relatedworks} discusses literature survey over closely related works. In Section \ref{sec:Network-Architecture-of-PALM},
The network architecture of both type-1
and type-2 PALM are elaborated. Section \ref{sec:Online-Learning-Policy_t1} describes
the online learning policy of type-1 PALM, while Section \ref{sec:Online-Learning-Policy_t2}
presents online learning mechanism of type-2 PALM. In Section \ref{sec:Evaluation},
the proposed PALM's efficacy has been evaluated through real-world
and synthetic data streams. Finally, the paper ends by drawing the
concluding remarks in Section \ref{sec:Conclusions}.

\section{Related Work and Research Gap With the State-of-The-Art Algorithms\label{sec:relatedworks}}

SANFS can be employed for data stream regression, since they can learn from scratch
with no base knowledge and are embedded with the self-organizing property
to adapt to the changing system dynamics \cite{lughofer2008flexfis}.
It fully work in a single-pass learning scenario, which is efficient for online learning under limited computational resources. An early work in this domain is seen in \cite{juang1998online} where an SANFS, namely SONFIN, was proposed. Evolving clustering method (ECM) is implemented
in \cite{kasabov2002denfis} to evolve fuzzy rules. Another pioneering work in this area is
the development of the online evolving T-S fuzzy system namely eTS
\cite{angelov2004approach} by Angelov. eTS has been improved in the several follow-up works: eTS+ \cite{angelov2010evolving},
Simpl\_eTS \cite{angelov2005simpl_ets}, AnYa \cite{angelov2012new}.
However, eTS+, and Simpl\_eTS generate axis parallel ellipsoidal clusters,
which cannot deal effectively with non-axis parallel data distribution.
To deal with the non-axis parallel data distribution, an evolving
multi-variable Gaussian (eMG) function was introduced in the fuzzy
system in \cite{lemos2013adaptive}. Another example of SANFS exploiting the multivarible Gaussian function is found in \cite{pratama2014panfis} where the concept of statistical contribution is implemented to grow and prune the fuzzy rules on the fly. This work has been extended in \cite{pratama2014genefis} where the idea of statistical contribution is used as a basis of input contribution estimation for the online feature selection scenario.

The idea of SANFS was implemented in type-2 fuzzy system in \cite{juang2008self}.
Afterward, they have extended their concept in local recurrent architecture
\cite{juang2010recurrent}, and interactive recurrent architecture
\cite{juang2013data}. These works utilize Karnik-Mendel (KM) type reduction technique \cite{karnik1999type}, which relies on an iterative approach to find left-most and right-most points. To mitigate
this shortcoming, the KM type reduction technique can be replaced with
$q$ design coefficient \cite{watkins1992q} introduced
in \cite{abiyev2010type}. SANFS is also introduced under the context of metacognitive learning machine (McLM) which encompasses three fundamental pillars of human learning: what-to-learn, how-to-learn, when-to-learn. The idea of McLM was
introduced in \cite{suresh2010sequential}. McLM has been modified with the use of Scaffolding theory, McSLM, which aims to realize the plug-and-play learning fashion \cite{pratama2016incremental}. To solve
the problem of uncertainty, temporal system dynamics and the unknown
system order McSLM was extended in recurrent
interval-valued metacognitive scaffolding fuzzy neural network (RIVMcSFNN)
\cite{pratama2017data}. The vast majority of SANFSs are developed using the concept of HSSC and HESC which impose considerable memory demand and computational burden because both rule premise and rule consequent have to be stored and evolved during the training process.

\section{Network Architecture of PALM\label{sec:Network-Architecture-of-PALM}}

In this section, the network architecture of PALM is presented in details. The T-S
fuzzy system is a commonly used technique to approximate complex nonlinear
systems due to its universal approximation property. The
rule base in the T-S fuzzy model of that multi-input single-output
(MISO) system can be expressed in the following IF-THEN rule
format:

\begin{align}
R^{j}:\quad & \text{If}\;x_{1}\;\text{is}\;B_{1}^{j}\;\text{and}\;x_{2}\;\text{is}\;B_{2}^{j}\;\text{and}...\text{and}\;x_{n}\;\text{is}\;B_{n}^{j}\;\nonumber \\
 & \text{Then}\;y_{j}=b_{0j}+a_{1j}x_{1}+...+a_{nj}x_{n}\label{eq:fuz_R}
\end{align}
where $R^{j}$ stands for the $j$th rule, $j=1,2,3,...,R,$ and $R$
indicates the number of rules, $i=1,2,...,n;$ $n$ denotes the dimension of
input feature, $x_{n}$ is the $n$th input feature, $a$ and b are
consequent parameters of the sub-model belonging to the $j$th rule,
$y_{j}$ is the output of the $j$th sub-model. The T-S fuzzy model
can approximate a nonlinear system with a combination of several piecewise linear
systems by partitioning the entire input space into several
fuzzy regions. It expresses each input-output space with a linear
equation as presented in (\ref{eq:fuz_R}). Approximation using T-S
fuzzy model leads to a nonlinear programming problem and hinders its
practical use. A simple solution to the problem is the utilization
of various clustering techniques to identify the rule premise parameters. Because of the generation of the
linear equation in the consequent part, the HPBC can be applied to construct the T-S fuzzy system efficiently. The advantages of using HPBC in the T-S
fuzzy model can be seen graphically in Fig. \ref{fig:Clustering-in-T-S}.

\begin{center}
\begin{figure}[tb]
\begin{centering}
\includegraphics[scale=0.41]{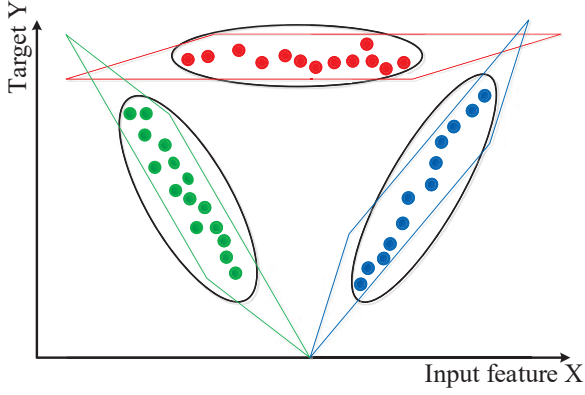}
\par\end{centering}
\caption{Clustering in T-S fuzzy model using hyperplanes\label{fig:Clustering-in-T-S}}
\end{figure}
\par\end{center}

Some popular algorithms with HPBC are fuzzy C-regression model (FCRM)
\cite{hathaway1993switching}, fuzzy C-quadratic shell (FCQS) \cite{krishnapuram1995fuzzy},
double FCM \cite{kim1997new}, inter type-2 fuzzy c-regression model
(IT2-FCRM) \cite{zou2017ts}. A main limitation of these algorithms is
their non-incremental nature which does not suit for data stream regression. Moreover, they still deploy Gaussian function to represent the rule premise of TS fuzzy model which does not exploit the parameter efficiency trait of HPBC. To fill up this research gap, a new membership function \cite{zou2017ts}
is proposed to accommodate the use of hyperplanes in the rule premise part of TS fuzzy system. It
can be expressed as:

\begin{equation}
\mu_{B}(j)=\exp\left(-\Gamma\frac{dst(j)}{\max\left(dst(j)\right)}\right)\label{eq:MF}
\end{equation}

where $j=1,2,...,R;$ $R$ is the number of rules, $\Gamma$ is an adjustment parameter which controls the fuzziness of membership grades. Based on the observation in \cite{zou2017ts}, and empirical analysis with variety of data streams in our work, the range of $\Gamma$ is settled as $[1, 100]$.
	$dst(j)$ denotes the distance from present sample to the
	$j$th hyperplane. In our work, $dst(j)$ is
	defined as \cite{zou2017ts} as follows:

\begin{equation}
dst(j)=\frac{|X_t\omega_j|}{||\omega_j||}\label{eq:dis_p_hp}
\end{equation}
where $X_t\in\Re^{1\times(n+1)}$ and $\omega_j\in\Re^{(n+1)\times1}$ respectively stand for the input vector of the $t$th observation and the output weight vector of the $j$th rule. This membership function enables the incorporation of HPBC directly into the T-S fuzzy system directly with the absence of rule parameters except the first order linear function or hyperplane. Because a point to plane distance is not unique, the compatibility measure is executed using the minimum point to plane distance. The following discusses the network structure of PALM encompassing its type-1 and type-2 versions. PALM can be modeled as a four-layered network working in tandem, where the fuzzy rule triggers a hyperplane-shaped cluster and is induced by \eqref{eq:dis_p_hp}. Since T-S fuzzy rules can be developed solely using a hyperplane, PALM
is free from antecedent parameters which results in dramatic reduction of network parameters. Furthermore, it operates in the one-pass learning fashion where it works point by point and a data point is discarded directly once learned.

\subsection{Structure of Type-1 PALM Network:}

In type-1 PALM network architecture, the membership function exposed in (\ref{eq:MF}) is utilized to fit the hyperplane-shaped cluster in identifying type-1 T-S fuzzy model. To understand the work flow let us consider that a single data point $x_{n}$ is fed into PALM at the $n-th$ observation. Appertaining to the concept of type-1 fuzzy system, this crisp data needs to be transformed to fuzzy set. This fuzzification process is attained using  type-1 hyperplane-shaped membership function, which is framed through the concept of point-to-plane distance. This hyperplane-shaped type-1 membership function can be expressed as:

\begin{equation}
f_{T1}^{1}=\mu_{B}(j)=\exp\left(-\Gamma\frac{dst(j)}{\max\left(dst(j)\right)}\right)\label{eq:mu_T1}
\end{equation}
where $dst(j)$ in \eqref{eq:mu_T1} denotes the distance between the current sample and $j$th hyperplane as with \eqref{eq:dis_p_hp}. It is defined as per definition of a point-to-plane distance \cite{pointPlane} and is formally expressed as follows:

\begin{equation}
dst(j)=\bigg|\frac{y_{d}-(\sum_{i=1}^{n}a_{ij}x_{i}+b_{0j})}{\sqrt{1+\sum_{i=1}^{n}(a_{ij})^{2}}}\bigg|\label{eq:dis_p_hp-t2}
\end{equation}

where $a_{ij}$ and $b_{0j}$ are consequent parameters of the $j$th rule, $i=1,2,...,n;$
	$n$ is the number of input dimension, and $y_d$ is the target variable. The exertion of $y_d$ is an obstruction for PALM due to target variable's unavailability in testing phase. This issue comes into picture due to the definition of a point-to-hyperplane distance \cite{pointPlane}. To eradicate such impediment, a recurrent PALM (RPALM) framework is developed here. We refer curious readers to the supplementary document for details on the RPALM. Considering a MISO system, the IF-THEN rule of type-1 PALM can be expressed as follows:

\begin{align}
R^{j}:\quad & \text{IF}\;X_{n}\;\text{is close to}\;f_{T1_{j}}^{2}\;\text{THEN}\;y_{j}=x_{e}^{j}\omega_{j}\label{eq:fuz_R-1}
\end{align}
where $x_{e}$ is the extended input vector and is
expressed  by inserting the intercept to the original input vector as $x_{e}=[1,x_{1}^{k},x_{2}^{k},...,x_{n}^{k}]$, $\omega_{j}$
is the weight vector for the $j$th rule, $y_{j}$ is the consequent
part of the $j$th rule. Since type-1 PALM has no premise parameters, the antecedent part is simply hyperplane. It is observed from (\ref{eq:fuz_R-1}) that the drawback of HPBC-based TS fuzzy system lies in the high level fuzzy inference scheme which degrades the transparency of fuzzy rule. The intercept of extended input vector controls the slope of hyperplane which functions to prevent the untypical gradient problem.

The consequent part is akin to the basic T-S fuzzy model's rule
consequent part $(y_{j}=b_{0j}+a_{1j}x_{1}+...+a_{nj}x_{n})$. The
consequent part for the $j$th hyperplane is calculated
by weighting the extended input variable $(x_{e})$ with its corresponding weight vector as follows:

\begin{equation}
f_{T1_{j}}^{2}=x_{e}^{T}\omega_{j}\label{eq:consequent}
\end{equation}
It is used in (\ref{eq:consequent}) after updating recursively by the
FWGRLS method, which ensures a smooth change in the weight value. In the next step, the rule firing strength is normalized and combined with the rule consequent to produce the end-output of type-1 PALM. The final crisp output of the PALM for type-1 model can be expressed as follows:

\begin{equation}
	f_{T1}^{3}=\frac{\sum_{j=1}^{R}f_{T1_{j}}^{1}f_{T1_{j}}^{2}}{\sum_{i=1}^{R}f_{T1_{i}}^{1}}\label{eq:y_l}
	\end{equation}
The normalization term in (\ref{eq:y_l}) guarantees the partition of unity where the sum of normalized membership degree is unity. The T-S fuzzy system is functionally-equivalent to the radial basis function (RBF) network if the rule firing strength is directly connected to the output of the consequent layer \cite{wu2000dynamic}. It is also depicted that the final crisp output is produced by the weighted average defuzzification scheme.

\subsection{Network structure of the Type-2 PALM :}

Type-2 PALM differs from the type-1 variant in the use of interval-valued hyperplane generating the type-2 fuzzy rule. Akin to its type-1 version, type-2 PALM starts operating by intaking the crisp input data stream $x_{n}$ to be fuzzied. Here, the fuzzification occurs with help of interval-valued hyperplane based membership function, which can be expressed as:

\begin{equation}
\widetilde{f}_{out}^{1}=\exp\left(-\Gamma\frac{\widetilde{dst}(j)}{\max\left(\widetilde{dst}(j)\right)}\right)
\end{equation}
where $\widetilde{f}_{out}^{1}=\left[\underline{f}_{out}^{1},\;\overline{f}_{out}^{1}\right]$ is the upper and lower hyperplane, $\widetilde{dst}(j)=\left[\overline{dst}(j),\;\underline{dst}(j)\right]$ is interval valued distance, where $\overline{dst}(j)$ is the distance between present input samples and $j$th upper hyperplane, and $\underline{dst}(j)$ is that between present input samples and $j$th lower hyperplane. In type-2 architecture, distances among incoming input data and upper and lower hyperplanes are calculated as follows:

\begin{equation}
	\widetilde{dst}(j)=\bigg|\frac{y_{d}-(\sum_{i=1}^{n}\widetilde{a}_{ij}x_{i}+\widetilde{b}_{0j})}{\sqrt{1+\sum_{i=1}^{n}(\widetilde{a}_{ij})^{2}}}\bigg|\label{eq:dis_p_hp-t2-1}
	\end{equation}
where $\widetilde{a}_{ij}=\left[\underline{a}_{ij};\overline{a}_{ij}\right]$ and $\widetilde{b}_{0j}=\left[\underline{b}_{0j};\overline{b}_{0j}\right]$ are the interval-valued coefficients of the rule consequent of type-2 PALM. Like the type-1 variants, type-2 PALM has dependency on target value ($y_d$). Therefore, they are also extended into type-2 recurrent structure and elaborated in the supplementary document. The use of interval-valued coefficients result in the interval-valued firing strength which forms the footprint of uncertainty (FoU). The FoU is the key component against uncertainty of data streams and sets the degree of tolerance against uncertainty.

In a MISO system, the IF-THEN rule of type-2 PALM can be
expressed as:

\begin{align}
	R^{j}:\quad & \text{IF}\;X_{n}\;\text{is close to}\;\widetilde{f}_{out}^{2}\;\text{THEN}\;y_{j}=x_{e}^{j}\widetilde{\omega}_{j}\label{eq:fuz_R-1-1}
	\end{align}
where $x_{e}$ is the extended input vector, $\widetilde{\omega}_{j}$
is the interval-valued weight vector for the $j$th rule, $y_{j}$ is the consequent
part of the $j$th rule, whereas the antecedent part is merely interval-valued hyperplane. The type-2 fuzzy rule is similar to that of the type-1 variant except the presence of interval-valued firing strength and interval-valued weight vector. In type-2 PALM, the consequent part is calculated by weighting the extended input variable $x_e$ with the interval-valued output weight vectors $\widetilde{\omega_j}=\left[\underline{\omega_j},~\overline{\omega_j}\right]$ as follows:

\begin{equation}
	\overline{f}_{out_{j}}^{2}=x_{e}^{j}\overline{\omega}_{j},\quad\underline{f}_{out_{j}}^{2}=x_{e}^{j}\underline{\omega}_{j}\label{eq:consequent-1}
	\end{equation}
The lower weight vector $\underline{\omega}_j$ for the $j$th lower hyperplane, and upper weight vector $\overline{\omega}_j$ for the $j$th upper hyperplane are initialized by allocating higher value for upper weight vector than the lower weight vector. These vectors are updated recursively by FWGRLS method, which ensures a smooth change in weight value.

Before performing the defuzzification method, the type reduction mechanism is carried out to craft the type-reduced set - the transformation from the type-2 fuzzy variable to the type-1 fuzzy variable. One of the commonly used type-reduction method is the Karnik Mendel (KM)
procedure \cite{karnik1999type}. However, in the KM method, there is an involvement
of an iterative process due to the requirement of reordering
the rule consequent first in ascending order before getting the cross-over
points iteratively incurring expensive computational cost. Therefore, instead of the KM method, the $q$ design
factor \cite{watkins1992q} is utilized to orchestrate the type reduction process. The final crisp output of the type-2 PALM
can be expressed as follows:

\begin{equation}
	f_{out}^{3}=y_{out}=\frac{1}{2}\left(y_{l_{out}}+y_{r_{out}}\right)\label{eq:y-1}
	\end{equation}
where

\begin{equation}
	y_{l_{out}}=\frac{\sum_{j=1}^{R}q_{l}\underline{f}_{out}^{1}\underline{f}_{out}^{2}}{\sum_{i=1}^{R}\overline{f}_{out}^{1}}+\frac{\sum_{j=1}^{R}(1-q_{l})\overline{f}_{out}^{1}\underline{f}_{out}^{2}}{\sum_{i=1}^{R}\underline{f}_{out}^{1}}\label{eq:y_l-1}
	\end{equation}

\begin{equation}
	y_{r_{out}}=\frac{\sum_{j=1}^{R}q_{r}\underline{f}_{out}^{1}\overline{f}_{out}^{2}}{\sum_{i=1}^{R}\overline{f}_{out}^{1}}+\frac{\sum_{j=1}^{R}(1-q_{r})\overline{f}_{out}^{1}\overline{f}_{out}^{2}}{\sum_{i=1}^{R}\underline{f}_{out}^{1}}\label{eq:y_r-1}
	\end{equation}
where $y_{l_{out}}$ and $y_{r_{out}}$ are the left and right outputs resulted from the type reduction mechanism. $q_{l}$ and $q_{r}$, utilized
in (\ref{eq:y_l-1}) and (\ref{eq:y_r-1}), are the design factors initialized in a way to satisfy
the condition $q_{l}<q_{r}$. In our $q$ design factor, the $q_{l}$ and $q_{r}$ steers the proportion
of the upper and lower rules to the final crisp outputs $y_{l_{out}}$
and $y_{r_{out}}$ of the PALM. The normalization process of the type-2 fuzzy inference scheme \cite{abiyev2010type} was modified in
\cite{pratama2017data} to prevent the generation of the invalid interval.
The generation of this invalid interval as a result of the normalization process
of \cite{abiyev2010type} was also proved in \cite{pratama2017data}.
Therefore, normalization process as adopted in \cite{pratama2017data}
is applied and advanced in terms of $q_{l}$ and $q_{r}$ in our
work. Besides, in order to improve the performance of the proposed
PALM, the $q_{l}$ and $q_{r}$ are not left constant rather continuously adapted using gradient
decent technique as explained in section \ref{sec:Online-Learning-Policy_t1}. Notwithstanding that the type-2 PALM is supposed to handle uncertainty better than its type-1 variant, it incurs a higher number of network parameters in the level of $2\times R \times(n+1)$ as a result of the use of upper and lower weight vectors $\widetilde{\omega_j}=\left[\underline{\omega}_j, \overline{\omega}_j\right]$. In addition, the implementation of q-design factor imposes extra computational cost because $q_{l}$ and $q_{r}$ call for a tuning procedure with the gradient descent method.

\section{Online Learning Policy in Type-1 PALM\label{sec:Online-Learning-Policy_t1}}

This section describes the online learning policy of our proposed
type-1 PALM. PALM is capable of starting its learning process from scratch with an empty rule base. Its fuzzy rules can be automatically generated on the fly using the self constructive clustering (SCC) method which checks the input and output coherence. The complexity reduction mechanism is implemented using the hyperplane merging module which vets similarity of two hyperplanes using the distance and angle concept. The hyperplane-based fuzzy rule is adjusted using the FWGRLS method in the single-pass learning fashion.

\subsection{Mechanism of Growing Rules}

The rule growing mechanism of type-1 PALM is adopted from the self-constructive
clustering (SSC) method developed in \cite{xu2015dimensionality,jiang2011fuzzy}
to adapt the number of rules. This method has been successfully applied to automatically generate interval-valued data clouds in \cite{pratama2018parsimonious} but its use for HPBC deserves an in-depth investigation. In this technique, the rule significance
is measured by calculating the input and output coherence. The coherence
is measured by analysing the correlation between the existing data
samples and the target concept. Hereby assuming the input vector as
$X_{t}\in\Re^{n}$, target vector as $T_{t}\in\Re^{n}$, hyperplane
of the $i$th local sub-model as $\mathcal{H}_{i}\in\Re^{1\times(n+1)}$,
the input and output coherence between $X_{t}\in\Re^{n}$ and each
$\mathcal{H}_{i}\in\Re^{1\times(n+1)}$ are calculated as follows:

\begin{equation}
I_{c}(\mathcal{H}_{i},\,X_{t})=\xi(\mathcal{H}_{i},\,X_{t})
\end{equation}

\begin{equation}
O_{c}(\mathcal{H}_{i},\,X_{t})=\xi(X_{t},T_{t})-\xi(\mathcal{H}_{i},T_{t})
\end{equation}
where $\xi(\,)$ express the correlation function. There are
various linear and nonlinear correlation methods for measuring correlation, which can be applied. Among them, the nonlinear methods
for measuring the correlation between variables are hard to employ
in the online environment since they commonly use the discretization
or Parzen window method. On the other hand, Pearson correlation is
a widely used method for measuring correlation between two variables.
However, it suffers from some limitations: it's insensitivity
to the scaling and translation of variables and sensitivity to rotation
\cite{mitra2002unsupervised}. To solve these problems, a method namely
maximal information compression index (MCI) is proposed in \cite{mitra2002unsupervised},
which has also been utilized in the SSC method to measure the correlation
$\xi(\,)$ between variables as follows:

\begin{align}
 & \xi(X_{t},T_{t})=\frac{1}{2}(\text{var}(X_{t})+\text{var}(T_{t})\nonumber \\
 & -\sqrt{\text{(var}(X_{t})+\text{var}(T_{t}))^{2}-4\text{var}(X_{t})(T_{t})(1-\rho(X_{t},T_{t})^{2})})\label{eq:correlation}
\end{align}

\begin{equation}
\rho(X_{t},T_{t})=\frac{\text{cov}(X_{t},T_{t})}{\sqrt{\text{var}(X_{t})\text{var}(T_{t})}}\label{eq:Pearson}
\end{equation}
where $\text{var}(X_{t})\text{,}\;\text{var}(T_{t})$ express
the variance of $X_{t}$ and $T_{t}$ respectively, $\text{cov}(X_{t},T_{t})$ presents the covariance between two variables $X_{t}$ and $T_{t}$,
$\rho(X_{t},T_{t})$ stands for Pearson correlation index of $X_{t}$
and $T_{t}$. In a similar way, the correlation $\xi(\mathcal{H}_{i},\,X_{t})$
and $\xi(\mathcal{H}_{i},T_{t})$ can be measured using (\ref{eq:correlation})
and (\ref{eq:Pearson}). In addition, the MCI method measures the
compressed information when a newly observed sample is ignored. Properties
of the MCI method in our work can be expressed as follows:
\begin{enumerate}
\item $0\leq\xi(X_{t},T_{t})\leq\frac{1}{2}(\text{var}(X_{t})+\text{var}(T_{t})).$
\item a maximum possible correlation is $\xi(X_{t},T_{t})=0$.
\item express symmetric behavior $\xi(X_{t},T_{t})=\xi(T_{t},X_{t})$.
\item invariance against the translation of the dataset.
\item express the robustness against rotation.
\end{enumerate}
$I_{c}(\mathcal{H}_{i},\,X_{t})$ is projected to explore the similarity
between $\mathcal{H}_{i}$ and $X_{t}$ directly, while $O_{c}(\mathcal{H}_{i},\,X_{t})$
is meant to examine the dissimilarity between $\mathcal{H}_{i}$
and $X_{t}$ indirectly by utilizing the target vector as a reference.
In the present hypothesis, the input and output coherence need to
satisfy the following conditions to add a new rule or hyperplane:

\begin{equation}
I_{c}(\mathcal{H}_{i},\,X_{t})>b_{1}\quad\text{and}\quad O_{c}(\mathcal{H}_{i},\,X_{t})<b_{2}\label{eq:cond_add_rul}
\end{equation}
where $b_{1}\in[0.01,\;0.1],$ and $b_{2}\in[0.01,\;0.1]$ are predetermined
thresholds. If the hypothesis satisfies both the conditions of (\ref{eq:cond_add_rul}),
a new rule is added with the highest input coherence. Besides, the
accommodated data points of a rule are updated as $N_{j^{*}}=N_{j^{*}}+1$.
Also, the correlation measure functions $\xi(\;)$ are updated with
(\ref{eq:correlation}) and (\ref{eq:Pearson}). Due to the utilization
of the local learning scenario, each rule is adapted separately and
therefore covariance matrix is independent to each rule $C_{j}(k)\in\Re^{(n+1)\times(n+1)}$,
here $n$ is the number of inputs. When a new hyperplane is added
by satisfying (\ref{eq:cond_add_rul}), the hyperplane parameters and  the output covariance matrix of FWGRLS method are crafted as follows:

\begin{equation}
\pi_{R+1}=\pi_{R^{*}},\quad C_{R+1}=\Omega I\label{eq:ini_of_Cj}
\end{equation}

Due to the utilization of the local learning scenario, the consequent
of the newly added rules can be assigned as the closest rule, since the
expected trend in the local region can be portrayed easily from the
nearest rule. The value of $\Omega$ in (\ref{eq:ini_of_Cj}) is very
large $(10^{5})$. The reason for initializing the C matrix with a
large value is to obtain a fast convergence to the real solution
\cite{ljung1999system}. The proof of such consequent parameter setting is detailed in \cite{lughofer2011evolving}. In addition,
the covariance matrix of the individual rule has no relationship with
each other. Thus, when the rules are pruned in the rule merging module,
the covariance matrix, and consequent parameters are deleted as it
does not affect the convergence characteristics of the C matrix and
consequent of remaining rules.

\subsection{Mechanism of Merging Rules}

In SANFS, the rule evolution mechanism usually generate
redundant rules. These unnecessary rules create complicacy in the
rule base, which hinders some
desirable features of fuzzy rules: transparency and tractability
in their operation. Notably, in handling data streams, two overlapping
clusters or rules may easily be obtained when new samples occupied
the gap between the existing two clusters. Several useful methods
have been employed to merge redundant rules or clusters in \cite{angelov2010evolving,lughofer2015generalized,pratama2014genefis,pratama2018parsimonious}.
However, all these techniques are appropriate for mainly hypersphere-based or ellipsoid-based
clusters.

In realm of hyperplane clusters, there is a possibility
of generating a higher number of hyperplanes in dealing with the same
dataset than spherical or ellipsoidal clusters because of the nature of HPBC in which each hyperplane represents specific operating region of the approximation curve. This opens higher chance in generating redundant rules than HSSC and HESC. Therefore, an appropriate merging technique is vital
and has to achieve tradeoff between diversity of fuzzy rules and generalization power of the rule base. To understand clearly, the merging of two hyperplanes
due to the new incoming training data samples is illustrated
in Fig. \ref{fig:Merging-of-redundant}.

\begin{center}
\begin{figure*}[tbh]
\begin{centering}
\includegraphics[scale=0.33]{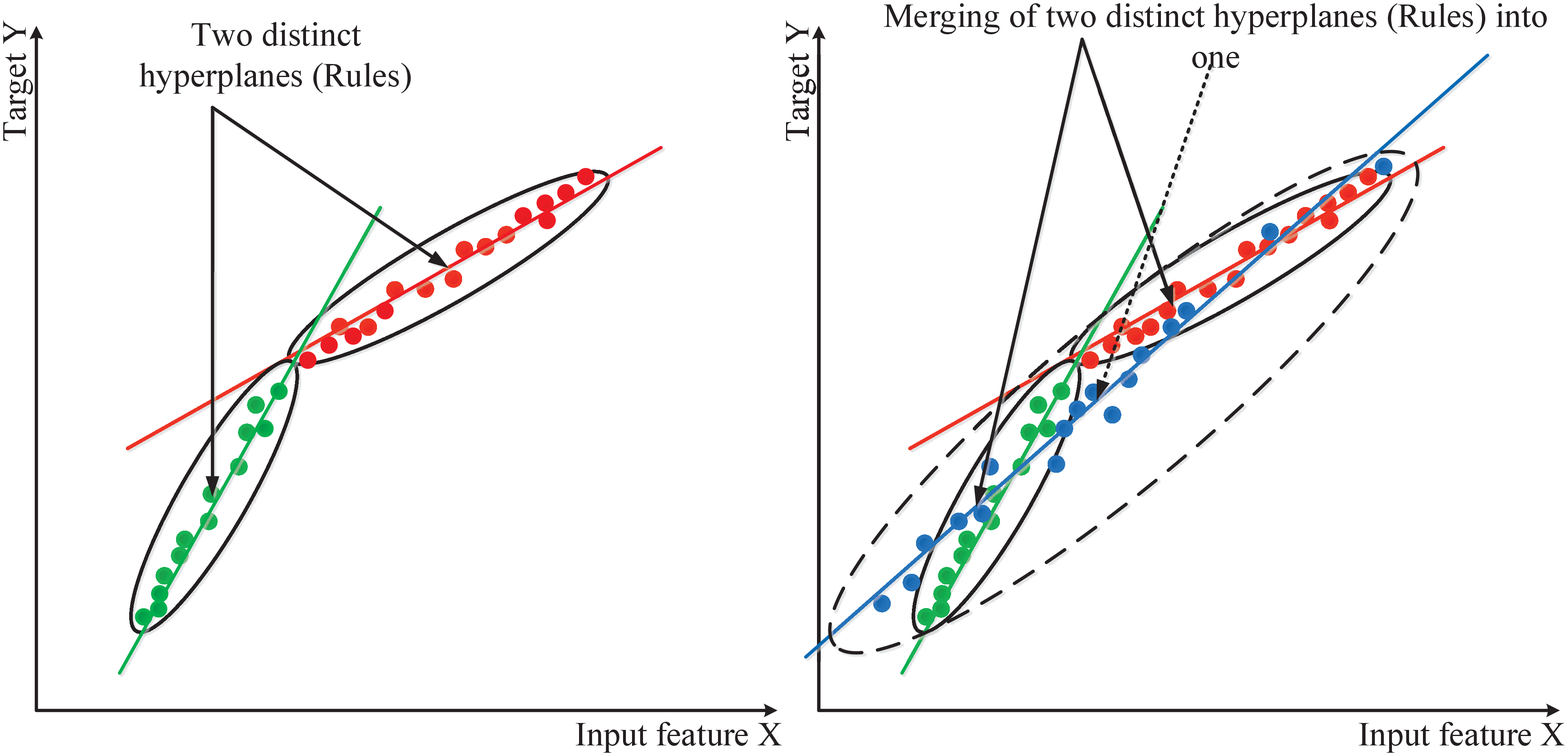}
\par\end{centering}
\caption{Merging of redundant hyperplanes (rules) due to newly incoming training
samples\label{fig:Merging-of-redundant}}

\end{figure*}
\par\end{center}

In \cite{kim2007incremental}, to merge the hyperplanes, the similarity
and dissimilarity between them are obtained by measuring only the
angle between the hyperplanes. This strategy is ,however, not conclusive to decide the similarity between two hyperplanes because it solely considers the orientation of hyperplane without looking at the relationship of two hyperplanes in the target space.

In our work, to measure the similarity between the hyperplane-shaped
fuzzy rules, the angle between them is estimated as follows \cite{lughofer2011line,pratama2014genefis}:

\begin{equation}
\theta_{hp}=\arccos\left(\bigg|\frac{\omega_{R}^{T}\omega_{R+1}}{|\omega_{R}||\omega_{R+1}|}\bigg|\right)\label{eq:angle_r_r1}
\end{equation}
where $\theta_{hp}$ is ranged between $0$ and $\pi$ radian, $\omega_{R}=\left[b_{1,R},\,b_{2,R},...,b_{k,R}\right],\;\omega_{R+1}=\left[b_{1,R+1},\,b_{2,R+1},...,b_{k,R+1}\right].$
The angle between the hyperplanes is not sufficient to decide whether the rule merging scenario should take place because it does not inform the closeness of two hyperplanes in the target space. Therefore, the spatial proximity between two hyperplanes in the hyperspace are taken into account. If we consider two hyperplanes as $l_{R1}=a_{1}+xb_{1},$
and $l_{R2}=a_{2}+xb_{2},$ then the minimum distance between them
can be projected as follows:

\begin{equation}
d_{R,R+1}=\bigg|(a_{1}-a_{2}).\frac{(b_{1}\times b_{2})}{|b_{1}\times b_{2}|}\bigg|\label{eq:dis_r_r1}
\end{equation}
The rule merging condition is formulated as follows:

\begin{equation}
\theta_{hp}\leq c_{1}\;\text{and}\;d_{R,R+1}\leq c_{2}\label{eq:condition_merge}
\end{equation}
where $c_{1}\in[0.01,0.1]$, $c_{2}\in[0.001,0.1]$ are predefined
thresholds. If (\ref{eq:condition_merge})
is satisfied, fuzzy rules are merged. It is worth noting that the merging technique is only
applicable in the local learning context because, in case of global learning,
the orientation and similarity of two hyperplanes have no direct correlation to their relationship.

In our merging mechanism, a dominant rule having higher support is retained, whereas a less dominant hyperplane (rule) resided by less number of samples is pruned to mitigate the structural simplification scenario of PALM. A dominant rule has a higher influence on the merged cluster because it represents the underlying data distribution. That is, the dominant rule is kept in the rule base in order for good partition of data space to be maintained and even improved. For simplicity, the weighted average strategy is adopted in merging two hyperplanes as follows:

\begin{equation}
\omega_{acm}^{new}=\frac{\omega_{acm}^{old}N_{acm}^{old}+\omega_{acm+1}^{old}N_{acm+1}^{old}}{N_{acm}^{old}+N_{acm+1}^{old}}\label{eq:merged_w}
\end{equation}

\begin{equation}
N_{acm}^{new}=N_{acm}^{old}+N_{acm+1}^{old}
\end{equation}
where $\omega_{acm}^{old}$ is the output weight vector of
the $acm$th rule, $\omega_{acm+1}^{old}$ is the output weight vector of $(acm+1)$th
rule, and $\omega_{acm}^{new}$ is the output weight vector of the merged rule,
$N$ is the population of a fuzzy rule. Note that the rule $acm$ is more influential than the rule $acm+1$,
since $N_{acm}>N_{acm+1}.$ The rule merging procedure is committed during the stable period where no addition of rules occurs. This strategy aims to attain a stable rule evolution and prevents new rules to be merged straightaway after being introduced in the rule base. As an alternative, the Yager's participatory learning-inspired merging scenario \cite{lughofer2015generalized} can be used to merge the two hyperplanes.

\subsection{Adaptation of Hyperplanes}

In previous work on hyperplane based T-S fuzzy system \cite{kim2006evolving}, recursive least
square (RLS) method is employed to calculate parameters of hyperplane.
As an advancement to the RLS method, a term for decaying the consequent parameter in the cost function
of the RLS method is utilized in \cite{xu2006generalized} and helps to obtain a solid generalization
performance - generalized recursive least square (GRLS) approach. However, their approach is formed in the context of global learning. A local learning method has some advantages over its global
counterpart: interpretability and robustness over noise. The interpretability is supported by the fact that each hyperplane portrays specific operating region of approximation curve. Also,
in local learning, the generation or deletion of any rule does not
harm the convergence of the consequent parameters of other rules,
which results in a significantly stable updating process \cite{angelov2008evolving}.

Due to the desired features of local learning scenario, the GRLS method
is extended in \cite{pratama2014genefis,pratama2017data}: Fuzzily Weighted Generalised Recursive Least Square (FWGRLS) method.
FWGRLS can be seen also as a variation of Fuzzily Weighted Recursive
Least Square (FWRLS) method \cite{angelov2004approach} with insertion of weight decay term. The FWGRLS method
is formed in the proposed type-1 PALM, where the cost function can
be expressed as:

\begin{align}
J_{L_{j}}^{n}= & (y_{t}-x_{e}\pi_{j})\Lambda_{j}(y_{t}-x_{e}\pi_{j})+\nonumber \\
 & 2\beta\varphi(\pi_{j})+(\pi-\pi_{j})(C_{j}x_{e})^{-1}(\pi-\pi_{j})
\end{align}

\begin{equation}
J_{L}^{n}=\sum_{j=1}^{i}J_{L_{j}}^{n}
\end{equation}
where $\Lambda_{j}$ denotes a diagonal matrix with the diagonal element
of $R_{j}$, $\beta$ represents a regularization parameter, $\varphi$
is a decaying factor, $x_{e}$ is the extended input vector,
$C_{j}$ is the covariance matrix, $\pi_{j}$ is the local subsystem
of the $j$th hyperplane. Following the similar approach as \cite{pratama2014genefis},
the final expression of the FWGRLS approach
is formed as follows:

\begin{align}
\pi_{j}(k)= & \pi_{j}(k-1)-\beta C_{j}(k)\nabla\varphi\pi_{j}(k-1)+\nonumber \\
 & \Upsilon(k)(y_{t}(k)-x_{e}\pi_{j}(k));\;j=[1,2,...,R]
\end{align}
where

\begin{flushright}
\begin{equation}
C_{j}(k)=C_{j}(k-1)-\Upsilon(k)x_{e}C_{j}(k-1)
\end{equation}
\par\end{flushright}

\begin{equation}
\Upsilon(k)=C_{j}(k-1)x_{e}\left(\frac{1}{\Lambda_{j}}+x_{e}C_{j}(k-1)x_{e}^{T}\right)^{-1}
\end{equation}
with the initial conditions

\begin{equation}
\pi_{1}(1)=0\quad\text{and}\quad C_{1}(1)=\Omega I
\end{equation}
where $\Upsilon(k)$ denotes the Kalman gain, $R$ is the number
of rules, $\Omega=10^{5}$ is a large positive constant. In this work,
the regularization parameter $\beta$ is assigned as an extremely
small value $(\beta\thickapprox10^{-7})$. It can be observed that the FWGRLS
method is similar to the RLS method without the term $\beta\pi_{j}(k)\nabla\varphi(k)$.
This term steers the value of $\pi_{j}(k)$ even to update an insignificant
amount of it minimizing the impact of inconsequential rules. The quadratic weight decay function is chosen in PALM written as follows:

\begin{equation}
\varphi(\pi_{j}(k-1))=\frac{1}{2}(\pi_{j}(k-1))^{2}
\end{equation}

Its gradient can be expressed as:

\begin{equation}
\nabla\varphi(\pi_{j}(k-1))=\pi_{j}(k-1)
\end{equation}

By utilizing this function, the adapted-weight is shrunk to a factor
proportional to the present value. It helps to intensify the generalization
capability by maintaining dynamic of output weights into small values
\cite{mackay1992bayesian}.

\section{Online Learning Policy in Type-2 PALM\label{sec:Online-Learning-Policy_t2}}

The learning policy of the type-1 PALM is extended in the context of the type-2
fuzzy system, where q design factor is utilized to carry out the type-reduction scenario.
The learning mechanisms are detailed in the following subsections.

\subsection{Mechanism of Growing Rules}

In realm of the type-2 fuzzy system, the SSC method has been extended
to the type-2 SSC (T2SSC) in \cite{pratama2018parsimonious}. It has been
adopted and extended in terms of the design factors $q_{l}$ and $q_{r}$,
since the original work in \cite{pratama2018parsimonious} only deals
with a single design factor $q$. In this T2SSC method,
the rule significance is measured by calculating the input and output
coherence as done in the type-1 system. By assuming $\mathcal{\widetilde{H}}_{i}=[\mathcal{\overline{H}}_{i},\mathcal{\underline{H}}_{i}]\in\Re^{R\times(1+n)}$
as interval-valued hyperplane of the $i$th local sub-model, the input
and output coherence for our proposed type-2 system can be extended
as follows:

\begin{equation}
I_{c_{L}}(\mathcal{\widetilde{H}}_{i},\,X_{t})=(1-q_{l})\xi(\mathcal{\overline{H}}_{i},\,X_{t})+q_{l}\xi(\mathcal{\underline{H}}_{i},\,X_{t})
\end{equation}

\begin{equation}
I_{c_{R}}(\mathcal{\widetilde{H}}_{i},\,X_{t})=(1-q_{r})\xi(\mathcal{\overline{H}}_{i},\,X_{t})+q_{r}\xi(\mathcal{\underline{H}}_{i},\,X_{t})
\end{equation}

\begin{equation}
I_{c}(\mathcal{\widetilde{H}}_{i},\,X_{t})=\frac{\left(I_{c_{L}}(\mathcal{\widetilde{H}}_{i},\,X_{t})+I_{c_{R}}(\mathcal{\widetilde{H}}_{i},\,X_{t})\right)}{2}\label{eq:Ic_t2}
\end{equation}

\begin{equation}
O_{c}(\mathcal{\widetilde{H}}_{i},\,X_{t})=\xi(X_{t},T_{t})-\xi(\mathcal{\widetilde{H}}_{i},T_{t})
\end{equation}

where

\begin{equation}
\xi_{L}(\mathcal{\widetilde{H}}_{i},T_{t})=(1-q_{l})\xi(\mathcal{\overline{H}}_{i},\,T_{t})+q_{l}\xi(\mathcal{\underline{H}}_{i},\,T_{t})
\end{equation}

\begin{equation}
\xi_{R}(\mathcal{\widetilde{H}}_{i},T_{t})=(1-q_{r})\xi(\mathcal{\overline{H}}_{i},\,T_{t})+q_{r}\xi(\mathcal{\underline{H}}_{i},\,T_{t})
\end{equation}

\begin{equation}
\xi(\mathcal{\widetilde{H}}_{i},T_{t})=\frac{\left(\xi_{L}(\mathcal{\widetilde{H}}_{i},T_{t})+\xi_{R}(\mathcal{\widetilde{H}}_{i},T_{t})\right)}{2}
\end{equation}

Unlike the direct calculation of input coherence $I_{c}(\;)$ in type-1
system, in type-2 system the $I_{c}(\;)$ is calculated using (\ref{eq:Ic_t2})
based on left $I_{c_{L}}(\;)$ and right $I_{c_{R}}(\;)$ input coherence.
By using the MCI method in the T2SCC rule growing process, the correlation is measured using (\ref{eq:correlation})
and (\ref{eq:Pearson}), where $(X_{t},T_{t})$ are substituted with
$(\mathcal{\overline{H}}_{i},\,X_{t})$ , $(\mathcal{\underline{H}}_{i},\,X_{t})$,
$(\mathcal{\overline{H}}_{i},\,T_{t})$ , $(\mathcal{\underline{H}}_{i},\,T_{t})$.
The conditions for growing rules remain the same as expressed in (\ref{eq:cond_add_rul}) and is only modified to fit the type-2 fuzzy system platform. The parameter settings for
the predefined thresholds are as with the type-1 fuzzy model.

\subsection{Mechanism of Merging Rules}

The merging mechanism of the type-1 PALM is extended for the
type-2 fuzzy model. To merge the rules, both the angle and distance between
two interval-valued hyperplanes are measured as follows:

\begin{equation}
\widetilde{\theta}_{hp}=\arccos\left(\bigg|\frac{\widetilde{\omega}_{R}^{T}\widetilde{\omega}_{R+1}}{|\widetilde{\omega}_{R}||\widetilde{\omega}_{R+1}|}\bigg|\right)
\end{equation}

\begin{equation}
\widetilde{d}_{R,R+1}=\bigg|(\widetilde{a}_{1}-\widetilde{a}_{2}).\frac{(\widetilde{b}_{1}\times\widetilde{b}_{2})}{|\widetilde{b}_{1}\times\widetilde{b}_{2}|}\bigg|
\end{equation}
where $\widetilde{\theta}_{hp}=[\overline{\theta}_{hp}\;\underline{\theta}_{hp}]$,
and $\widetilde{d}_{R,R+1}=[\overline{d}_{R,R+1}\;\underline{d}_{R,R+1}]$.
This $\widetilde{\theta}_{hp}$ and $\widetilde{d}_{R,R+1}$ also
needs to satisfy the condition of (\ref{eq:condition_merge}) to merge
the rules, where the same range of $c_{1}$ and $c_{2}$ are
applied in the type-2 PALM. The formula of merged weight in (\ref{eq:merged_w})
is extended for the interval-valued merged weight as follows:

\begin{equation}
\widetilde{\omega}_{acm}^{new}=\frac{\widetilde{\omega}_{acm}^{old}N_{acm}^{old}+\widetilde{\omega}_{acm+1}^{old}N_{acm+1}^{old}}{N_{acm}^{old}+N_{acm+1}^{old}}
\end{equation}
where $\widetilde{\omega}_{acm}=[\overline{\omega}_{acm}\;\underline{\omega}_{acm}]$. As with the type-1 PALM, the weighted average strategy is followed in the rule merging procedure of the type-2 PALM.

\subsection{Learning of the Hyperplane Submodels Parameters}

The FWGRLS method \cite{pratama2014genefis} is extended to
adjust the upper and lower hyperplanes
of the interval type-2 PALM. The final expression of the FWGRLS method is shown
as follows:

\begin{align}
\widetilde{\pi}_{j}(k)= & \widetilde{\pi}_{j}(k-1)-\beta\widetilde{C}_{j}(k)\nabla\varphi\widetilde{\pi}_{j}(k-1)+\nonumber \\
 & \widetilde{\Upsilon}(k)(y_{t}(k)-x_{e}\widetilde{\pi}_{j}(k));\;j=[1,2,...,R]
\end{align}
where

\begin{flushright}
\begin{equation}
\widetilde{C}_{j}(k)=\widetilde{C}_{j}(k-1)-\widetilde{\Upsilon}(k)x_{e}\widetilde{C}_{j}(k-1)
\end{equation}
\par\end{flushright}

\begin{equation}
	\widetilde{\Upsilon}(k)=\widetilde{C}_{j}(k-1)x_{e}\left(\frac{1}{\widetilde{\Lambda}_{j}}+x_{e}\widetilde{C}_{j}(k-1)x_{e}^{T}\right)^{-1}
	\end{equation}
where $\widetilde{\pi}_{j}=[\overline{\pi}_{j}\;\underline{\pi}_{j}]$,
$\widetilde{C}_{j}=[\overline{C}_{j}\;\underline{C}_{j}]$, $\widetilde{\Upsilon}=[\overline{\Upsilon}\;\underline{\Upsilon}]$,
and $\widetilde{\Lambda}_{j}=[\overline{\Lambda}_{j}\;\underline{\Lambda}_{j}]$.
The quadratic weight decay function of FWGRLS method remains in the type-2 PALM to provide the weight decay effect in the rule merging scenario.

\begin{center}
\begin{figure*}[t]
\begin{centering}
\subfloat[prediction]{\begin{centering}
\includegraphics[scale=0.17]{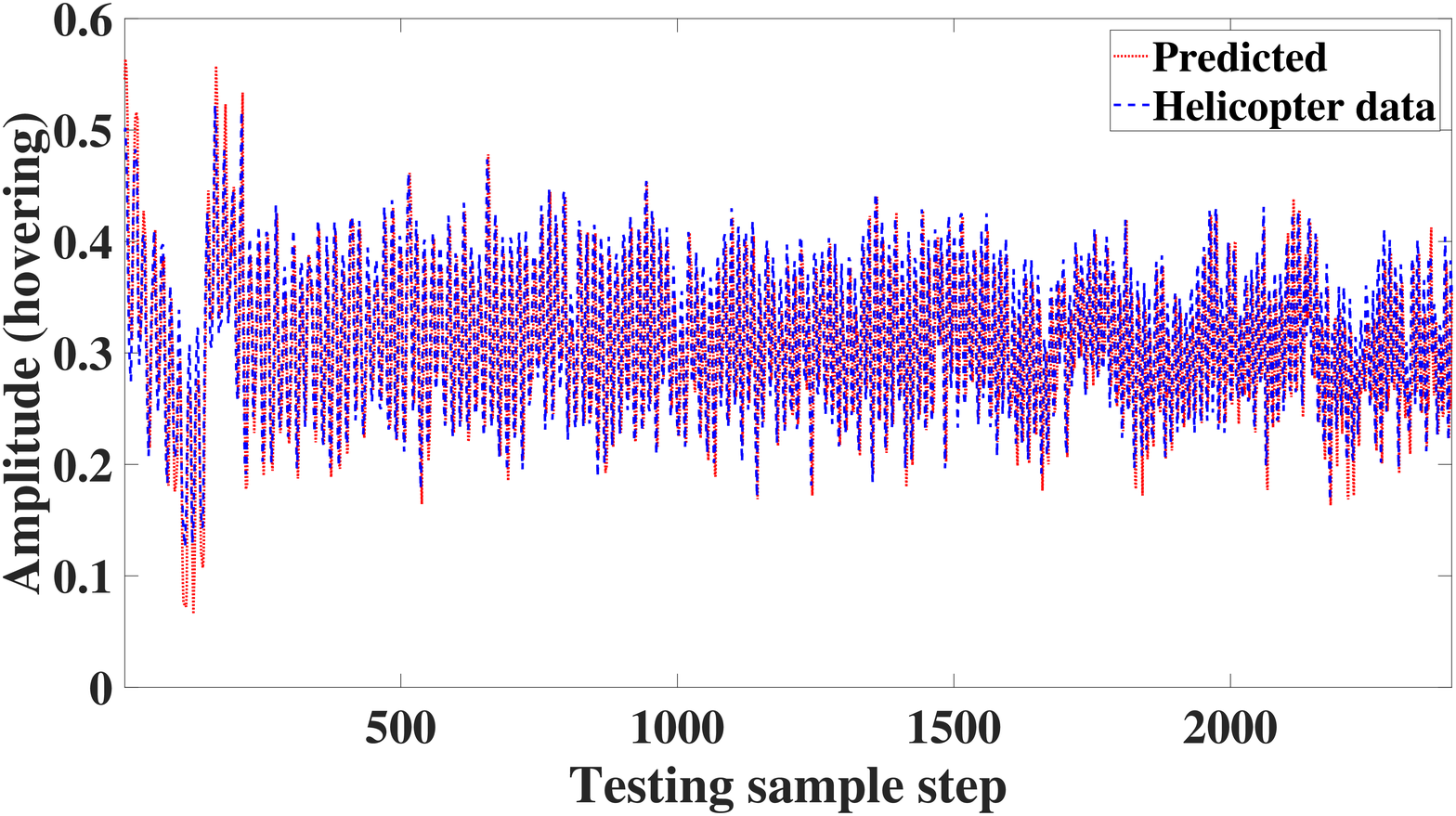}
\par\end{centering}
}\subfloat[rule evolution]{\begin{centering}
\includegraphics[scale=0.17]{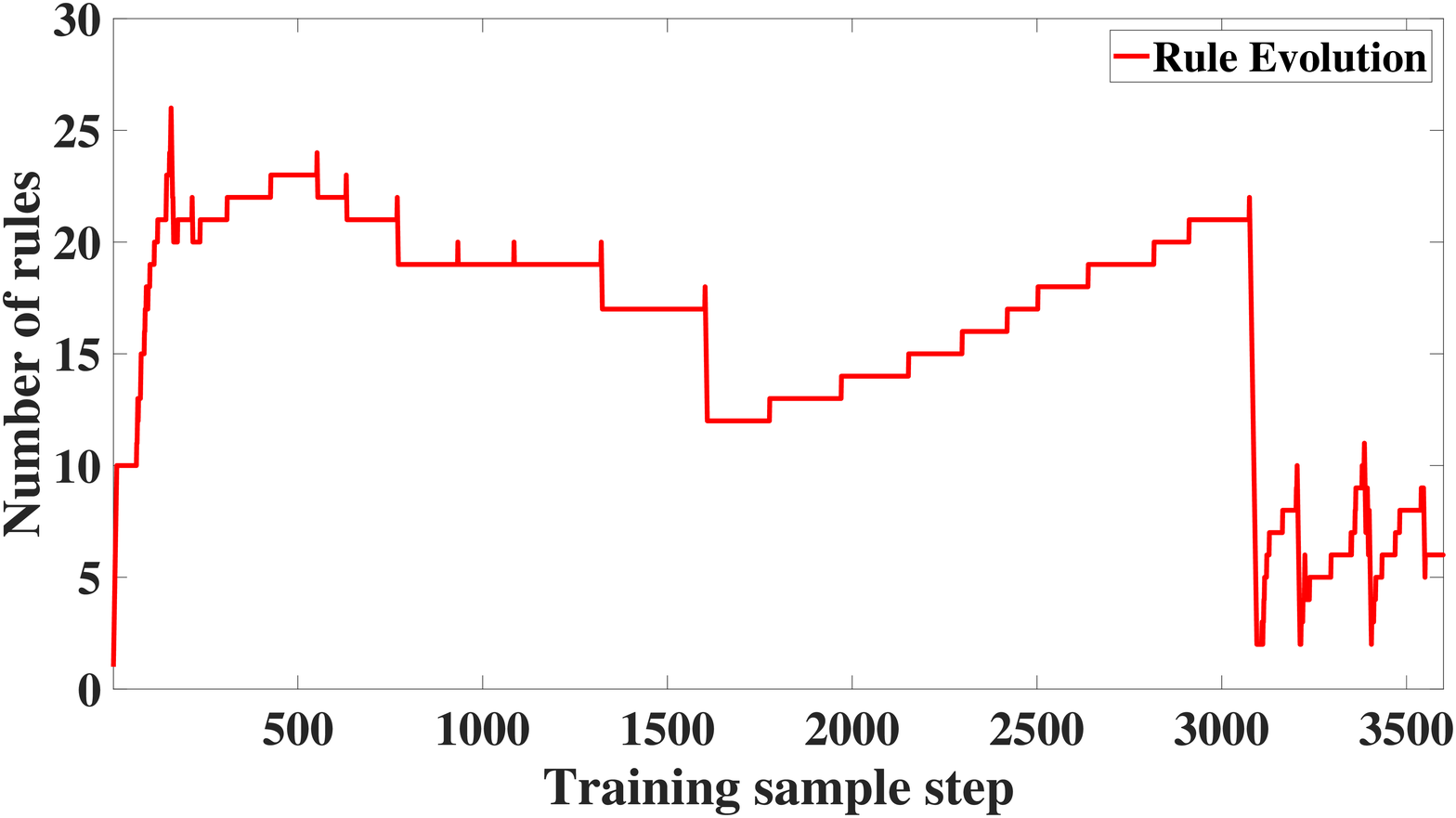}
\par\end{centering}

}
\par\end{centering}
\caption{(a) Online identification of helicopter (in hovering condition); (b)
rule evolution in that identification using type-2 PALM (L) \label{fig:(a)-Online-identification:heli} }

\end{figure*}
\par\end{center}

\subsection{Adaptation of $q$ Design Factors}

The $q$ design factor as used in \cite{pratama2017data}
is extended in terms of left $q_{l}$ and right $q_{r}$ design
factor to actualize a high degree of freedom of the type-2 fuzzy model. They are initialized in such a way that
the condition $q_{r}>q_{l}$ is maintained. In this adaptation process,
the gradient of $q_{l}$ and $q_{r}$ with respect to error $E=\frac{1}{2}\left(y_{d}-y_{out}\right)^{2}$
can be expressed as follows:

\begin{align}
\frac{\partial E}{\partial q_{l}} & =\frac{\partial E}{\partial y_{out}}\times\frac{\partial y_{out}}{\partial y_{l_{out}}}\times\frac{\partial y_{l_{out}}}{\partial q_{l}}\nonumber \\
 & =-\frac{1}{2}\left(y_{d}-y_{out}\right)\left(\frac{\underline{f}_{out}^{1}\underline{f}_{out}^{2}}{\sum_{i=1}^{R}\overline{f}_{out}^{1}}-\frac{\overline{f}_{out}^{1}\underline{f}_{out}^{2}}{\sum_{i=1}^{R}\underline{f}_{out}^{1}}\right)
\end{align}

\begin{align}
\frac{\partial E}{\partial q_{r}} & =\frac{\partial E}{\partial y_{out}}\times\frac{\partial y_{out}}{\partial y_{r_{out}}}\times\frac{\partial y_{r_{out}}}{\partial q_{r}}\nonumber \\
 & =-\frac{1}{2}\left(y_{d}-y_{out}\right)\left(\frac{\underline{f}_{out}^{1}\overline{f}_{out}^{2}}{\sum_{i=1}^{R}\overline{f}_{out}^{1}}-\frac{\overline{f}_{out}^{1}\overline{f}_{out}^{2}}{\sum_{i=1}^{R}\underline{f}_{out}^{1}}\right)
\end{align}

After obtaining the gradient $\frac{\partial E}{\partial q_{l}}$
and $\frac{\partial E}{\partial q_{r}}$, the $q_{l}$ and $q_{r}$
are updated using formulas as follows:

\begin{equation}
q_{l}^{new}=q_{l}^{old}-a\frac{\partial E}{\partial q_{l}^{old}}
\end{equation}

\begin{equation}
q_{r}^{new}=q_{r}^{old}-a\frac{\partial E}{\partial q_{r}^{old}}
\end{equation}
where $a=0.1$ is a learning rate. Note that the learning rate is a key of $q_{l}$ and $q_{r}$ convergence because it determines the step size of adjustment. An adaptive strategy as done in \cite{pratama2016incremental} can be implemented to shorten the convergence time without compromising the stability of adaptation process.  

\subsection{Impediments of the Basic PALM structure}
In the PALM, hyperplane-shaped membership function is formulated exercising a distance $(dst(j))$ exposed in \eqref{eq:dis_p_hp-t2}. The $(dst(j))$ is calculated using true output value based on theory of point to hyperplane distance \cite{pointPlane}. Therefore, the PALM has a dependency on the true output in deployment phase. Usually, true outputs are not known in the deployment mode. To circumvent such structural shortcoming, the so-called "Teacher Forcing" mechanism \cite{goodfellow2016deep} is employed in PALM. In teacher forcing technique, network has connections from outputs to their hidden nodes at the next time step. Based on this concept, the output of PALM is connected with the input layer at the next step, which constructs a recurrent PALM (RPALM) architecture. The modified distance formula for the RPALM architecture is provided in the supplementary document. Besides, the code of the proposed RPALM is made available in \cite{rpalmcode}. Our numerical results demonstrate that rPALM produces minor decrease of predictive accuracy compared to PALM but is still better than many of benchmarked SANFSs. The downside of the RPALM is that the rules are slightly not transparent because it relies on its predicted output of the previous time instant $y(k-1)$ rather than incoming input $x_k$.         

\begin{center}
\begin{table*}[tbh]
\caption{Modeling of the Box-Jenkins Time Series using various Self-Adaptive Neuro-Fuzzy Systems\label{tab:Modeling-of-the_BJ}}
\centering{}%
\begin{tabular}{|l|>{\centering}p{1.3cm}|>{\centering}p{1cm}|>{\centering}p{1cm}|>{\centering}p{1cm}|>{\centering}p{1.27cm}|>{\centering}p{1.27cm}|>{\centering}p{1.22cm}|>{\centering}p{1.35cm}|}
\hline 
Model  & Reference  & RMSE using testing samples & NDEI using testing samples & Number of rules & Number of inputs & Network Parameters & Number of training samples & Execution time (sec)\tabularnewline
\hline 
{DFNN } & {\cite{wu2000dynamic} } & {0.7800 } & {4.8619 } & {1 } & {2 } & {6 } & {200 } & {0.0933}\tabularnewline
\hline 
{GDFNN } & {\cite{wu2001fast} } & {0.0617 } & {0.3843 } & {1 } & {2 } & {7 } & {200 } & {0.0964}\tabularnewline
\hline 
FAOSPFNN & \cite{wang2009fast}& 0.0716 & 0.4466 & 1 &2 & 4  & {200 } & {0.0897}\tabularnewline
\hline 
{eTS } & {\cite{angelov2004approach} } & {0.0604 } & {0.3763 } & {5 } & {2 } & {30 } & {200 } & {0.0635}\tabularnewline
\hline 
{simp\_eTS } & {\cite{angelov2005simpl_ets} } & {0.0607 } & {0.3782 } & {3 } & {2 } & {18 } & {200 } & {1.5255}\tabularnewline
\hline 
{GENEFIS } & {\cite{pratama2014genefis} } & {0.0479 } & {0.2988 } & {2 } & {2 } & {18 } & {200 } & {0.0925}\tabularnewline
\hline 
{PANFIS } & {\cite{pratama2014panfis} } & {0.0672 } & {0.4191 } & {2 } & {2 } & {18 } & {200 } & {0.3162}\tabularnewline
\hline 
pRVFLN   & \cite{pratama2018parsimonious}   & 0.0478   & 0.2984   & 2   & 2   & 10   & 200   & 0.0614\tabularnewline
\hline 
Type-1 PALM (L)   & -   & 0.0484   & 0.3019   & 8   & 2   & 24   & 200   & 0.1972\tabularnewline
\hline 
Type-1 PALM (G)   & -   & 0.0439   & 0.2739   & 8   & 2   & 24   & 200   & 0.1244\tabularnewline
\hline 
Type-2 PALM (L)   & -   & 0.0377   & \textbf{0.2355}{}   & \textbf{2}{}   & 2   & 12   & 200   & 0.2723\tabularnewline
\hline 
Type-2 PALM (G)   & -   & \textbf{0.0066}{}   & \textbf{0.0410}{}   & 14   & 2   & 84   & 200   & 0.3558\tabularnewline
\hline 
\end{tabular}
\end{table*}
\par\end{center}

\section{Evaluation\label{sec:Evaluation}}

PALM has been evaluated through numerical studies with the use of synthetic ad real-world
streaming datasets. The code of PALMs and RPALMs along with these datasets have
been made publicly available in \cite{palmcode,rpalmcode}.

\subsection{Experimental Setup\label{subsec:Experimental-setup}}

\subsubsection{Synthetic Streaming Datasets}

Three synthetic streaming datasets are utilized in our work
to evaluate the adaptive mechanism of the PALM:
1) Box-Jenkins Time Series dataset; 2) the Mackey-Glass Chaotic Time
Series dataset; and 3) non-linear system identification dataset.

\paragraph{Box-Jenkins Gas Furnace Time Series Dataset}

The Box\textendash Jenkins (BJ) gas furnace dataset is a famous benchmark
problem in the literature to verify the performance of SANFSs. The objective of the BJ gas furnace problem
is to model the output $(y(k))$ i.e. the $CO_{2}$ concentration
from the time-delayed input $(u(k-4))$ methane flow rate and its previous output $y(k-1)$. The I/O configuration follows the standard setting in the literature as follows:

\begin{equation}
\widehat{y}(k)=f(u(k-4),y(k-1))\label{eq:BJ_data}
\end{equation}

This problem consists of 290 data samples (\ref{eq:BJ_data}) where 200 samples are reserved for the training samples while remaining 90 samples are used to test model's generalization.

\paragraph{Mackey-Glass Chaotic Time Series Dataset}

Mackey-Glass (MG) chaotic time series problem having its root in \cite{mackey1977oscillation}
is a popular benchmark problem to forecast the future
value of a chaotic differential delay equation by using the past values.
Many researchers have used the MG dataset to evaluate their SANFSs\textquoteright{}
learning and generalization performance. This dataset is characterized
by their nonlinear and chaotic behaviors where its nonlinear oscillations replicate most of the physiological processes. The MG
dataset is initially proposed as a control model of the generation
of white blood cells. The mathematical model is expressed as:

\begin{equation}
\frac{dy(k)}{dt}=\frac{by(k-\delta)}{1+y^{10}y(k-\delta)}-ay(k)
\end{equation}
where $b=0.2,$ $a=0.1,$ and $\delta=85$. The chaotic element is
primarily attributed by $\delta\geq17$. Data samples are generated through the fourth-order Range Kutta method and our goal is to predict
the system output $\widehat{y}(k+85)$ at $k=85$ using four inputs: $y(k),\;y(k-6),\;y(k-12)$, and $y(k-18)$. This
series-parallel regression model can be expressed as follows:

\begin{equation}
\widehat{y}(k+85)=f\left(y(k),\;y(k-6),\;y(k-12),\;y(k-18)\right)
\end{equation}

For the training purpose, a total of 3000 samples between $k=201$ and
$k=3200$ is generated with the help of the 4th-order Range-Kutta
method, whereas the predictive model is tested with unseen 500 samples in the
range of $k=5001-5500$ to assess the generalization capability
of the PALM.

\paragraph{Non-linear System Identification Dataset}

A non-linear system identification is put forward to validate the efficacy of PALM and has frequently been used by researchers to test their SANFSs. The nonlinear dynamic of the system can be formulated by the following differential equation:

\begin{equation}
y(k+1)=\frac{y(k)}{1+y^{2}(k)}+u^{3}(k)
\end{equation}
where $u(k)=\sin(2\pi k/100)$. The predicted output of the system
$\widehat{y}(k+1)$ depends on the previous inputs and its own lagged
outputs, which can be expressed as follows:

\begin{equation}
\widehat{y}(k+1)=f(y(k),y(k-1),...,y(k-10),u(k))
\end{equation}

The first 50000 samples are employed to build our predictive model, and other
200 samples are fed the model to test model's generalization.

\subsubsection{Real-World Streaming Datasets}

Three different real-world streaming datasets from two rotary wing
unmanned aerial vehicle's (RUAV) experimental flight tests and a time-varying
stock index forecasting data are exploited to study the performance of PALM.

\paragraph{Quadcopter Unmanned Aerial Vehicle Streaming Data}

A real-world streaming dataset is collected from a Pixhawk autopilot
framework based quadcopter RUAV's experimental flight test. All experiments
are performed in the indoor UAV laboratory at the University of New
South Wales, Canberra campus. To record quadcopter flight data, the
Robot Operating System (ROS), running under the Ubuntu 16.04 version
of Linux is used. By using the ROS, a well-structured communication
layer is introduced into the quadcopter reducing the burden of
having to reinvent necessary software.

During the real-time flight testing accurate vehicle position, velocity,
and orientation are the required information to identify the quadcopter
online. For system identification, a flight data of quadcopter's altitude
containing approximately 9000 samples are recorded with some noise
from VICON optical motion capture system. Among them, 60\% of the
samples are used for training and remaining 40\% are for testing.
In this work, our model's output $y(k)$ is estimated as $\widehat{y}(k)$
from the previous point $y(k-6),$ and the system input $u(k)$, which
is the required thrust to the rotors of the quadcopter. The regression
model from the quadcopter data stream can be expressed as follows:

\begin{equation}
\widehat{y}(k)=f\left(y(k-6),\;u(k)\right)
\end{equation}

\paragraph{Helicopter Unmanned Aerial Vehicle Streaming Data}

The chosen RUAV for gathering streaming dataset is a Taiwanese made
Align Trex450 Pro Direct Flight Control (DFC), fly bar-less, helicopter.
The high degree of non-linearity associated with the Trex450 RUAV
vertical dynamics makes it challenging to build a regression model
from experimental data streams. All experiments are conducted at
the UAV laboratory of the UNSW Canberra campus. Flight data consists
of 6000 samples collected in near hover, heave and in ground effect
flight conditions to simulate non-stationary environments. First 3600 samples are used for the training data,
and the rest of the data are aimed to test the model.
The nonlinear dependence of the helicopter RUAV is governed by the
regression model as follows:

\begin{equation}
\widehat{y}(k+1)=f\left(y(k),\;u(k)\right)
\end{equation}
where $\widehat{y}(k+1)$ is the estimated output of the helicopter
system at $k=1$.

\paragraph{Time-Varying Stock Index Forecasting Data}

Our proposed PALM has been evaluated by the time-varying dataset, namely the  prediction of Standard and
Poor\textquoteright s 500 (S\&P-500 (\textasciicircum{}GSPC)) market
index \cite{oentaryo2014online,tan2010bcm}. The dataset consists
of sixty years of daily index values ranging from 3 January 1950 to
12 March 2009, downloaded from \cite{financeyahooweb}. This problem comprises 14893 data samples. In our work,
the reversed order data points of the same 60 years indexes have amalgamated
with the original dataset, forming a new dataset with 29786
index values. Among them, 14893 samples are allocated to train the
model and the remainder of 14893 samples are used for the validation data. The target variable is the next day S\&P-500 index $y(k+1)$ predicted using previous five consecutive days indexes: $y(k),\;y(k-1),\;y(k-2)$,$\;y(k-3)$
and $y(k-4)$. The functional relationship of the predictive model is formalized as follows:

\begin{equation}
\widehat{y}(k+1)=f\left(y(k),\;y(k-1),\;y(k-2),\;y(k-3)\;y(k-4)\right)
\end{equation}

This dataset carries the sudden drift property which happens around 2008. This property corresponds to the economic recession in the US due to the housing crisis in 2009.

\begin{center}
\begin{table*}[tbh]
\caption{Modeling of the Mackey\textendash Glass Chaotic Time Series using
		various Self-Adaptive Neuro-Fuzzy Systems\label{tab:Modeling-of-the_MG}}
\centering{}%
\begin{tabular}{|l|>{\centering}p{1.3cm}|>{\centering}p{1cm}|>{\centering}p{1cm}|>{\centering}p{1cm}|>{\centering}p{1.27cm}|>{\centering}p{1.27cm}|>{\centering}p{1.22cm}|>{\centering}p{1.35cm}|}
\hline 
Model   & Reference   & RMSE using testing samples   & NDEI using testing samples   & Number of rules   & Number of inputs   & Network Parameters   & Number of training samples   & Execution time (sec)\tabularnewline
\hline 
DFNN   & \cite{wu2000dynamic}   & 3.0531   & 12.0463   & 1   & 4   & 10   & 3000   & 11.1674\tabularnewline
\hline 
GDFNN   & \cite{wu2001fast}   & 0.1520   & 0.6030   & 1   & 4   & 13   & 3000   & 12.1076\tabularnewline
\hline 
FAOSPFNN   & \cite{wang2009fast}   & 0.2360   & 0.9314   & 1   & 4   & 6   & 3000   & 13.2213\tabularnewline
\hline 
eTS   & \cite{angelov2004approach}   & 0.0734   & 0.2899   & 48   & 4   & 480   & 3000   & 8.6174\tabularnewline
\hline 
simp\_eTS   & \cite{angelov2005simpl_ets}   & 0.0623   & 0.2461   & 75   & 4   & 750   & 3000   & 20.9274\tabularnewline
\hline 
GENEFIS   & \cite{pratama2014genefis}   & 0.0303   & \textbf{0.1198}{}   & 42   & 4   & 1050   & 3000   & 4.9694\tabularnewline
\hline 
PANFIS   & \cite{pratama2014panfis}   & 0.0721   & 0.2847   & 33   & 4   & 825   & 3000   & 4.8679\tabularnewline
\hline 
pRVFLN   & \cite{pratama2018parsimonious}   & 0.1168   & 0.4615   & 2   & 4   & 18   & 2993   & 0.9236\tabularnewline
\hline 
Type-1 PALM (L)   & -   & 0.0688   & 0.2718   & 5   & 4   & 25   & 3000   & 0.8316\tabularnewline
\hline 
Type-1 PALM (G)   & -   & 0.0349   & 0.1380   & 18   & 4   & 90   & 3000   & \textbf{0.7771}\tabularnewline
\hline 
Type-2 PALM (L)   & -   & 0.0444   & 0.1755   & 11   & 4   & 110   & 3000   & 2.8138\tabularnewline
\hline 
Type-2 PALM (G)   & -   & \textbf{0.0159}{}   & \textbf{0.0685}{}   & 13   & 4   & 130   & 3000   & 2.4502\tabularnewline
\hline 
\end{tabular}
\end{table*}
\par\end{center}

\subsection{Results and Discussion}

In this work, we have developed PALM by implementing type-1 and type-2
fuzzy concept, where both of them are simulated under
two parameter optimization scenarios: 1) Type-1
PALM (L); 2) Type-1 PALM (G); 3) Type-2 PALM (L); 4) Type-2 PALM (G). $L$ denotes the $Local$ update strategy while $G$ stands for the $Global$ learning mechanism. Basic PALM models are tested with three synthetic and three real-world streaming datasets. Furthermore, the models are compared against eight prominent variants of SANFSs, namely DFNN \cite{wu2000dynamic}, GDFNN \cite{wu2001fast}, FAOSPFNN \cite{wang2009fast}, eTS \cite{angelov2004approach}, simp\_eTS \cite{angelov2005simpl_ets}, GENEFIS \cite{pratama2014genefis}, PANFIS \cite{pratama2014panfis}, and pRVFLN \cite{pratama2018parsimonious}. Experiments with real-world and synthesis data streams are repeated with recurrent PALM. All experimental results using the RPALM are also purveyed in the supplementary document. Proposed PALMs' efficacy has been evaluated by measuring the root mean square error (RMSE), and nondimensional error index (NDEI) written as follows:

\begin{equation}
MSE=\frac{\sum_{k=1}^{N}(y_{t}-y_{k})^{2}}{N_{T_{s}}},\;RMSE=\sqrt{MSE}
\end{equation}

\begin{equation}
NDEI=\frac{RMSE}{Std(T_{s})}
\end{equation}
where $N_{T_{s}}$ is the total number of testing samples, and $Std(T_{s})$
denotes a standard deviation over all actual output values in the
testing set. A comparison is produced under the same computational platform in Intel(R) Xeon(R) E5-1630 v4 CPU with a 3.70 GHz processor and 16.0 GB installed memory.

\begin{center}
\begin{table*}[tbh]
\caption{Modeling of the non-linear system using various Self-Adaptive Neuro-Fuzzy Systems\label{tab:Modeling-of-the_NDI}}
\centering{}%
\begin{tabular}{|l|>{\centering}p{1.3cm}|>{\centering}p{1cm}|>{\centering}p{1cm}|>{\centering}p{1cm}|>{\centering}p{1.27cm}|>{\centering}p{1.27cm}|>{\centering}p{1.22cm}|>{\centering}p{1.35cm}|}
\hline 
Model   & Reference   & RMSE using testing samples   & NDEI using testing samples   & Number of rules   & Number of inputs   & Network Parameters   & Number of training samples   & Execution time (sec)\tabularnewline
\hline 
DFNN   & \cite{wu2000dynamic}   & 0.0380   & 0.0404   & 2   & 2   & 12   & 50000   & 2149.246\tabularnewline
\hline 
GDFNN   & \cite{wu2001fast}   & 0.0440   & 0.0468   & 2   & 2   & 14   & 50000   & 2355.726\tabularnewline
\hline 
FAOSPFNN   & \cite{wang2009fast}   & 0.0027   & 0.0029   & 4   & 2   & 16   & 50000   & 387.7890\tabularnewline
\hline 
eTS   & \cite{angelov2004approach}   & 0.07570  & 0.08054  & 7   & 2   & 42   & 50000   & 108.5791\tabularnewline
\hline 
simp\_eTS  & \cite{angelov2005simpl_ets}  & 0.07417  & 0.07892  & 7   & 2   & 42   & 50000   & 129.5552\tabularnewline
\hline 
GENEFIS   & \cite{pratama2014genefis}  & \textbf{0.00041}  & \textbf{0.00043} & 6  & 2   & 54   & 50000  & 10.9021\tabularnewline
\hline 
PANFIS   & \cite{pratama2014panfis}   & 0.00264  & 0.00281  & 27   & 2   & 243   & 50000   & 42.4945\tabularnewline
\hline 
pRVFLN   & \cite{pratama2018parsimonious}   & 0.06395  & 0.06596  & 2   & 2   & 10   & 49999   & 12.0105\tabularnewline
\hline 
Type-1 PALM (L)   & -   & 0.08808  & 0.09371  & 5   & 2   & 15   & 50000   & \textbf{9.9177}\tabularnewline
\hline 
Type-1 PALM (G)   & -   & 0.07457  & 0.07804  & 9   & 2   & 27   & 50000   & 10.5712\tabularnewline
\hline 
Type-2 PALM (L)   & -   & 0.03277  & 0.03487  & 3   & 2   & 18   & 50000   & 13.7455\tabularnewline
\hline 
Type-2 PALM (G)   & -   & 0.00387  & 0.00412  & 21   & 2   & 126   & 50000   & 55.4865\tabularnewline
\hline 
\end{tabular}
\end{table*}
\par\end{center}

\begin{center}
\begin{table*}[tbh]
\caption{Online modeling of the quadcopter utilizing various Self-Adaptive Neuro-Fuzzy Systems\label{tab:Online-modeling-of_quadcopter}}
\centering{}%
\begin{tabular}{|l|>{\centering}p{1.3cm}|>{\centering}p{1cm}|>{\centering}p{1cm}|>{\centering}p{1cm}|>{\centering}p{1.27cm}|>{\centering}p{1.27cm}|>{\centering}p{1.22cm}|>{\centering}p{1.35cm}|}
\hline 
Model   & Reference   & RMSE using testing samples   & NDEI using testing samples   & Number of rules   & Number of inputs   & Network Parameters   & Number of training samples   & Execution time (sec)\tabularnewline
\hline 
DFNN   & \cite{wu2000dynamic}   & 0.1469   & 0.6925   & 1   & 2   & 6   & 5467   & 19.0962\tabularnewline
\hline 
GDFNN   & \cite{wu2001fast}   & 0.1442   & 0.6800   & 1   & 2   & 7   & 5467   & 20.1737\tabularnewline
\hline 
FAOSPFNN   & \cite{wang2009fast}   & 0.2141   & 1.0097   & 12   & 2   & 48   & 5467   & 25.4000\tabularnewline
\hline 
eTS   & \cite{angelov2004approach}   & 0.1361   & 0.6417   & 4   & 2   & 24   & 5467   & 3.0686\tabularnewline
\hline 
simp\_eTS   & \cite{angelov2005simpl_ets}   & 0.1282   & 0.6048   & 4   & 2   & 24   & 5467   & 3.9984\tabularnewline
\hline 
GENEFIS   & \cite{pratama2014genefis}   & 0.1327   & 0.6257   & 1   & 2   & 9   & 5467   & 1.7368\tabularnewline
\hline 
PANFIS   & \cite{pratama2014panfis}   & 0.1925   & 0.9077   & 47   & 2   & 424   & 5467   & 6.0244\tabularnewline
\hline 
pRVFLN   & \cite{pratama2018parsimonious}   & 0.1191   & 0.5223   & 1   & 2   & 5   & 5461   & 0.9485\tabularnewline
\hline 
Type-1 PALM (L)   & -   & 0.1311   & 0.6182   & 2   & 2   & 6   & 5467   & 0.6605\tabularnewline
\hline 
Type-1 PALM (G)   & -   & 0.1122   & 0.5290   & 2   & 2   & 6   & 5467   & \textbf{0.5161}\tabularnewline
\hline 
Type-2 PALM (L)   & -   & \textbf{0.1001}{}   & \textbf{0.4723}{}   & 3   & 2   & 18   & 5467   & 1.7049\tabularnewline
\hline 
Type-2 PALM (G)   & -   & \textbf{0.0326}{}   & \textbf{0.1538}{}   & 4   & 2   & 24   & 5467   & 1.6802\tabularnewline
\hline 
\end{tabular}
\end{table*}
\par\end{center}

\subsubsection{Results and Discussion on Synthetic Streaming Datasets}

Table \ref{tab:Modeling-of-the_BJ} sums up
the outcomes of the Box-Jenkins time series for all benchmarked models.
Among various models, our proposed type-2 PALM (G) clearly outperforms other consolidated algorithms
in terms of predictive accuracy. For instance, the measured NDEI is just
0.0598 - the lowest among all models. Type-2 PALM (G) generates
thirteen (13) rules to achieve this accuracy level. Although
the number of generated rules is higher than that of remaining models, this accuracy far exceeds its counterparts whose accuracy hovers around 0.29
A fair comparison is also established by utilizing very close
number of rules in some benchmarked strategies namely eTS, simp\_eTS,
PANFIS, and GENEFIS. By doing so the lowest observed NDEI among the
benchmarked variations is 0.29, delivered by GENEFIS. It is substantially
higher than the measured NDEI of type-2 PALM (G). The advantage of HPBC is evidenced by the number of PALM's network parameters, where with thirteen rules
and two inputs, PALM evolves only 39 parameters, whereas the number
of network parameters of other algorithm for instance GENEFIS is 117. PALM requires the fewest parameters
than all the other variants of SANFS as well and affects positively to execution speed of PALM. On the other hand, with
only one rule the NDEI of PALM is also lower than the benchmarked
variants as observed in type-2 PALM (L) from Table \ref{tab:Modeling-of-the_BJ},
where it requires only 3 network parameter. It is important to note
that the rule merging mechanism is active in the case of only local learning
scenario. Here the number of induced rules are 8 and 2, which is lower than
i.e. 8 and 14 in their global learning versions. In both cases
of G and L, the NDEI is very close to each other with a very similar number
of rules. In short, PALM constructs a compact regression model using
the Box-Jenkins time series with the least number of network parameters
while producing the most reliable prediction.

The prediction of Mackey\textendash Glass chaotic time series is
challenging due to the nonlinear and chaotic behavior. Numerical results on the Mackey\textendash Glass
chaotic time series dataset is consolidated in Table
\ref{tab:Modeling-of-the_MG}, where 500 unseen samples are used
to test all the models. Due to the highly nonlinear behavior, an NDEI lower than 0.2 was obtained
from only GENEFIS \cite{pratama2014genefis} among other benchmarked algorithms. However, it costs 42
rules and requires a big number (1050) of network parameters. On the
contrary, with only 13 rules, 65 network parameters and faster execution,
the type-2 PALM (G) attains NDEI of 0.0685,
where this result is traced within 2.45 seconds due to
the deployment of fewer parameters than its counterparts. The use of rule merging method in local learning mode
reduces the generated rules to five (5) - type-1 PALM (L).
A comparable accuracy is obtained from type-1 PALM (L) with only 5
rules and 25 network parameters. An accomplishment of such accuracy
with few parameters decreases the computational complexity in predicting
complex nonlinear system as witnessed from type-1 PALM (L) in Table
\ref{tab:Modeling-of-the_MG}. Due to low computational burden,
the lowest execution time of 0.7771 seconds is achieved by the type-1 PALM
(G).

PALM has been utilized to estimate a high-dimensional non-linear
system with 50000 training samples. this study case depicts similar trend where PALM is capable of delivering comparable accuracy but with much less computational complexity and memory demand. The deployment
of rule merging module lessens the number of rules from 9 to 5 in
case of type-1 PALM, and 3 from 21 in type-2 PALM. The obtained
NDEI of PALMs with such a small number of rules is also similar to
other SANFS variants. To sum up, the PALM can deal with streaming examples with low computational burden due to
the utilization of few network parameters, where it maintains a
comparable or better predictive accuracy.

\subsubsection{Results and Discussion on Real-World Data Streams}

Table \ref{tab:Online-modeling-of_quadcopter} outlines the results
of online identification of a quadcopter RUAV from experimental flight
test data. A total 9112 samples of quadcopter's hovering test with
a very high noise from motion capture technique namely VICON \cite{vicon}
is recorded. Building SANFS using the noisy
streaming dataset is computationally expensive as seen from a high
execution time of the benchmarked SANFSs. Contrast with these standard
SANFSs, a quick execution time is seen from PALMs. It happens
due to the requirement of few network parameters. Besides, PALM arrives at encouraging accuracy as well. For instance,
the lowest NDEI at just 0.1538 is elicited in type-2 PALM (G).
To put it plainly, due to utilizing incremental HPBC, PALM can perform
better than its counterparts SANFSs driven by HSSC and HESC methods when dealing
with noisy datasets.

The online identification of a helicopter
RUAV (Trex450 Pro) from experimental flight data at hovering condition
are tabulated in Table \ref{tab:Online-modeling-of_heli}. The highest
identification accuracy with the NDEI of only 0.1380 is
obtained from the proposed type-2 PALM (G) with 9 rules. As with the previous experiments, the activation of rule merging
scenario reduces the fuzzy rules significantly from 11 to 6 in type-1
PALM, and from 9 to 6 in type-2 PALM. The highest accuracy
is produced by type-2 PALM with only 4 rules due to most likely uncertainty
handling capacity of type-2 fuzzy system. PALM's prediction on the helicopter's hovering dynamic and its rule evolution are depicted in Fig. \ref{fig:(a)-Online-identification:heli}. These figures are produced by the type-2 PALM(L). For further clarification, the fuzzy rule extracted by type-1 PALM(L) in case of modeling helicopter can be uttered as follows:
	\begin{align}\label{eq:T1rule_eg}
	&R^{1}: \text{IF}~X~\text{is close to}~\bigg([1,x_{1},x_{2}]\times\\\nonumber
	&[0.0186,-0.0909,0.9997]^{T}\bigg),~\text{THEN}~y_{1}=0.0186-0.0909x_{1}\\\nonumber
	&+0.9997x_{2}
	\end{align}
	In \eqref{eq:T1rule_eg}, the antecedent part is manifesting the hyperplane. The consequent part is simply $y_{1}=x_{e}^{1}\omega,$ where $\omega \in\Re ^{(n+1)\times 1}$, $n$ is the number of input dimension. Usage of 2 inputs in the experiment of Table V assembles an extended input vector
	like: $x_{e}^{1}=[1,x_{1},x_{2}]$. The weight
	vector is: $\left[\omega_{01},\omega_{11},\omega_{21}\right]=[0.3334,0.3334,0.3334]$ 
	In case of Type-2 local learning configuration, a rule can be stated as follows:
	\begin{align}\label{eq:T2rule_eg}
	&R^{1}: \text{IF}~X~\text{is close to}~\bigg(\big([1,x_{1},x_{2}]\times\\\nonumber
	&[0.0787,-0.3179,1.0281]^{T}\big),([1,x_{1},x_{2}]\times\\\nonumber
	&[0.2587,-0.1767,1.2042]^{T})\bigg)\\\nonumber
	&\text{THEN}~y_{1}=[0.0787,0.2587]+[-0.3179,-0.1767]x_{1}+\\\nonumber
	&[1.0281,1.2042]x_{2}
	\end{align}
	where (\ref{eq:T2rule_eg}) is expressing the first rule among
	6 rules formed in that experiment in Type-2 PALM's local learning scenario. Since
	the PALM has no premise parameters, the antecedent part is just presenting
	the interval-valued hyperplanes. The consequent part is noting but
	$y_{1}=x_{e}^{1}\widetilde{\omega}$, where $\widetilde{\omega} \in\Re ^{(2(n+1))\times 1}$, $n$ is the number of input dimension. Since 2 inputs are availed
	in the experiment of Table V, the extended input vector
	is: $x_{e}^{1}=[1,x_{1},x_{2}]$, and interval-valued weight
	vectors are: $\left[\underline{\omega_{01}},~\overline{\omega_{01}}\right]=[0.0787,0.2587];$
	$\left[\underline{\omega_{11}},~\overline{\omega_{11}}\right]=[-0.3179,-0.1767];$
	$\left[\underline{\omega_{21}},~\overline{\omega_{21}}\right]=[1.0281,1.2042].$
Furthermore, the predictive capability, rule evolution, NDEI evolution
and error of the PALM for six streaming datasets are attached in the supplementary document to keep the paper compact.

The numerical results on the time-varying Stock Index Forecasting S\&P-500 (\textasciicircum{}GSPC) problem are organized in Table \ref{tab:Modeling-of-the_SP500}. The lowest number of network parameters is obtained from PALMs, and subsequently, the fastest training speed of 2.0326 seconds is attained by type-1 PALM (L). All consolidated benchmarked algorithms generate the same level of accuracy around 0.015 to 0.06.

\begin{center}
\begin{table*}[tbh]
\caption{Online modeling of the helicopter utilizing various Self-Adaptive Neuro-Fuzzy Systems\label{tab:Online-modeling-of_heli}}
\centering{}%
\begin{tabular}{|l|>{\centering}p{1.3cm}|>{\centering}p{1cm}|>{\centering}p{1cm}|>{\centering}p{1cm}|>{\centering}p{1.27cm}|>{\centering}p{1.27cm}|>{\centering}p{1.22cm}|>{\centering}p{1.35cm}|}
\hline 
Model   & Reference   & RMSE using testing samples   & NDEI using testing samples   & Number of rules   & Number of inputs   & Network Parameters   & Number of training samples   & Execution time (sec)\tabularnewline
\hline 
DFNN   & \cite{wu2000dynamic}   & 0.0426   & 0.6644   & 1   & 2   & 6   & 3600   & 8.7760\tabularnewline
\hline 
GDFNN   & \cite{wu2001fast}   & 0.0326   & 0.5082   & 2   & 2   & 14   & 3600   & 11.2705\tabularnewline
\hline 
FAOSPFNN   & \cite{wang2009fast}   & 0.0368   & 0.5733   & 2   & 2   & 8   & 3600   & 2.4266\tabularnewline
\hline 
eTS   & \cite{angelov2004approach}   & 0.0535   & 0.8352   & 3   & 2   & 18   & 3600   & 1.3822\tabularnewline
\hline 
simp\_eTS   & \cite{angelov2005simpl_ets}   & 0.0534   & 0.8336   & 3   & 2   & 18   & 3600   & 2.3144\tabularnewline
\hline 
GENEFIS   & \cite{pratama2014genefis}   & 0.0355   & 0.5541   & 2   & 2   & 18   & 3600   & 0.6736\tabularnewline
\hline 
PANFIS   & \cite{pratama2014panfis}   & 0.0362   & 0.5652   & 9   & 2   & 81   & 3600   & 1.4571\tabularnewline
\hline 
pRVFLN   & \cite{pratama2018parsimonious}   & 0.0329   & 0.5137   & 2   & 2   & 10   & 3362   & 1.0195\tabularnewline
\hline 
Type-1 PALM (L)   & -   & 0.0363   & 0.5668   & 6   & 2   & 18   & 3600   & 0.9789\tabularnewline
\hline 
Type-1 PALM (G)   & -   & 0.0313   & 0.4886   & 11   & 2   & 33   & 3600   & 0.9517\tabularnewline
\hline 
Type-2 PALM (L)   & -   & 0.0201   & 0.3141   & 6   & 2   & 36   & 3600   & 2.3187\tabularnewline
\hline 
Type-2 PALM (G)   & -   & 0.0088   & \textbf{0.1380}{}   & 9   & 2   & 54   & 3600   & 1.9496\tabularnewline
\hline 
\end{tabular}
\end{table*}
\par\end{center}

\begin{center}
\begin{table*}[tbh]
\caption{Modeling of the Time-varying Stock Index Forecasting using various Self-Adaptive Neuro-Fuzzy Systems\label{tab:Modeling-of-the_SP500}}
\centering{}%
\begin{tabular}{|l|>{\centering}p{1.3cm}|>{\centering}p{1cm}|>{\centering}p{1cm}|>{\centering}p{1cm}|>{\centering}p{1.27cm}|>{\centering}p{1.27cm}|>{\centering}p{1.22cm}|>{\centering}p{1.35cm}|}
\hline 
Model   & Reference   & RMSE using testing samples   & NDEI using testing samples   & Number of rules   & Number of inputs   & Network Parameters   & Number of training samples   & Execution time (sec)\tabularnewline
\hline 
DFNN   & \cite{wu2000dynamic}   & 0.00441  & 0.01554  & 1   & 5   & 12   & 14893   & 347.7522\tabularnewline
\hline 
GDFNN   & \cite{wu2001fast}   & 0.30363  & 1.07075  & 1   & 5   & 16   & 14893   & 344.4558\tabularnewline
\hline 
FAOSPFNN   & \cite{wang2009fast}   & 0.20232  & 0.71346  & 1   & 5   & 7   & 14893   & 15.1439\tabularnewline
\hline 
eTS   & \cite{angelov2004approach}   & 0.01879  & 0.06629  & 3   & 5   & 36   & 14893   & 30.1606\tabularnewline
\hline 
simp\_eTS   & \cite{angelov2005simpl_ets}   & 0.00602  & 0.02124  & 3   & 5   & 36   & 14893   & 29.4296\tabularnewline
\hline 
GENEFIS   & \cite{pratama2014genefis}   & 0.00849  & 0.02994  & 3   & 5   & 108   & 14893   & 2.2076\tabularnewline
\hline 
PANFIS   & \cite{pratama2014panfis}   & 0.00464  & 0.01637  & 8   & 5   & 288   & 14893   & 5.2529\tabularnewline
\hline 
pRVFLN   & \cite{pratama2018parsimonious}   & 0.00441  & 0.01555  & 1   & 5   & 11   & 11170   & 2.5104\tabularnewline
\hline 
Type-1 PALM (L)   & -   & 0.00273  & 0.00964  & 3   & 5   & 18   & 14893   & \textbf{2.0326}\tabularnewline
\hline 
Type-1 PALM (G)   & -   & \textbf{0.00235} & \textbf{0.00832} & 5   & 5   & 30   & 14893   & 2.2802\tabularnewline
\hline 
Type-2 PALM (L)   & -   & 0.00442  & 0.01560  & 2   & 5   & 24   & 14893   & 4.0038\tabularnewline
\hline 
Type-2 PALM (G)   & -   & 0.00421  & 0.01487  & 3   & 5   & 36   & 14893   & 3.9134\tabularnewline
\hline 
\end{tabular}
\end{table*}
\par\end{center}

\subsection{Sensitivity Analysis of Predefined Thresholds}

In the rule growing purpose, two predefined thresholds ($b_{1}$ and $b_{2}$)
are utilized in our work. During various experimentation, it has been
observed that the higher the value of $b_{1}$, the less the number of hyperplanes
are added and vice versa. Unlike the effect of $b_{1}$, in case of
$b_{2}$, at higher values, more hyperplanes are added and vice versa.
To further validate this feature, the sensitivity of $b_{1}$ and
$b_{2}$ is evaluated using the Box\textendash Jenkins (BJ) gas furnace
dataset. The same I/O relationship as described in the subsection
\ref{subsec:Experimental-setup} is applied here, where the model
is trained also with same 200 samples and remaining 90 unseen samples
are used to test the model.

In the first test, $b_{2}$ is varied in the range of $[0.052,0.053,0.054,0.055],$
while the value of $b_{1}$ is kept fixed at 0.020. On the other hand,
the varied range for $b_{1}$ is $[0.020,0.022,0.024,0.026]$, while
$b_{2}$ is maintained at 0.055. In the second test, the altering
range for $b_{1}$ is $[0.031,0.033,0.035,0.037]$, and for $b_{2}$
is $[0.044,0.046,0.048,0.050].$ In this test, for a varying $b_{1}$,
the constant value of $b_{2}$ is 0.050, where $b_{1}$ is fixed
at 0.035 during the change of $b_{2}$. To evaluate the sensitivity
of these thresholds, normalized RMSE (NRMSE), NDEI, running time,
and number of rules are reported in Table \ref{tab:Sensitivity-Analysis}.
The NRMSE formula can be expressed as:

\begin{equation}
NRMSE=\sqrt{\frac{MSE}{Std(T_{s})}}
\end{equation}

From Table \ref{tab:Sensitivity-Analysis}, it has been observed that
in the first test for different values of $b_{1}$ and$b_{2}$, the
value of NRMSE and NDEI remains stable at 0.023 and 0.059 respectively.
The execution time varies in a stable range of $[0.31,0.35]$
seconds and the number of generated rules is 13. In the second test, the
NRMSE, NDEI, and execution time are relatively constant in the range of $[0.046,0.048]$,
$[0.115,0.121]$, $[0.26,0.31]$ correspondingly. The value of $b_{1}$ increases, and $b_{2}$ reduces compared to test 1, and the fewer
number of rules are generated across different experiments of our work.

\begin{table}[tbh]
\caption{Sensitivity Analysis of Rule growing thresholds\label{tab:Sensitivity-Analysis}}
\centering{}%
\begin{tabular}{|c|c|c|c|c|}
\hline
Parameters & NRMSE & NDEI & Execution time & \#Rules\tabularnewline
\hline
$b_{2}=0.055$ & 0.023 & 0.059 & 0.355 & 13\tabularnewline
\hline
$b_{2}=0.054$ & 0.023 & 0.059 & 0.312 & 13\tabularnewline
\hline
$b_{2}=0.053$ & 0.023 & 0.059 & 0.326 & 13\tabularnewline
\hline
$b_{2}=0.052$ & 0.023 & 0.059 & 0.325 & 13\tabularnewline
\hline
$b_{1}=0.020$ & 0.023 & 0.059 & 0.324 & 13\tabularnewline
\hline
$1_{2}=0.022$ & 0.023 & 0.059 & 0.325 & 13\tabularnewline
\hline
$b_{1}=0.024$ & 0.023 & 0.059 & 0.320 & 13\tabularnewline
\hline
$b_{1}=0.026$ & 0.023 & 0.059 & 0.344 & 13\tabularnewline
\hline
\hline
$b_{1}=0.037$ & 0.046 & 0.115 & 0.260 & 10\tabularnewline
\hline
$b_{1}=0.035$ & 0.046 & 0.115 & 0.259 & 11\tabularnewline
\hline
$b_{1}=0.033$ & 0.046 & 0.115 & 0.269 & 11\tabularnewline
\hline
$b_{1}=0.031$ & 0.048 & 0.121 & 0.269 & 11\tabularnewline
\hline
$b_{2}=0.050$ & 0.047 & 0.118 & 0.265 & 11\tabularnewline
\hline
$b_{2}=0.048$ & 0.046 & 0.115 & 0.267 & 11\tabularnewline
\hline
$b_{2}=0.046$ & 0.047 & 0.116 & 0.266 & 11\tabularnewline
\hline
$b_{2}=0.044$ & 0.047 & 0.117 & 0.306 & 11\tabularnewline
\hline
\end{tabular}
\end{table}

\section{Conclusions\label{sec:Conclusions}}

A novel SANFS, namely PALM, is proposed in this paper for data stream regression.
The PALM is developed with the concept of HPBC which incurs very low network parameters. The reduction of network parameters bring down the execution times because only the output weight vector calls for the tuning scenario without compromise on predictive accuracy. PALM possesses a highly adaptive rule base where its fuzzy rules can be automatically added when necessary based on the SCC theory. It implements the rule merging scenario for complexity reduction and the concept of distance and angle is introduced to coalesce similar rules. The efficiency of the PALM has been tested in six real-world and artificial data stream regression problems where PALM outperforms recently published works in terms of network parameters and running time. It also delivers state-of-the art accuracies which happen to be comparable and often better than its counterparts. In the future, PALM will be incorporated under a deep network structure.

\section*{Acknowledgment}

The authors would like to thank the Unmanned Aerial Vehicle laboratory
of the UNSW at the Australian Defense Force Academy for supporting
with the real-world datasets from the quadcopter and helicopter flight
test, and Computational Intelligence Laboratory of Nanyang Technological
University (NTU) Singapore for the computational support. This research
is financially supported by NTU start-up grant and MOE Tier-1 grant.

\bibliographystyle{IEEEtran}
\bibliography{refPALM}

\end{document}